%% file: main.tex
\documentclass[11pt]{article}

\usepackage[final]{acl}

\usepackage{times}
\usepackage{latexsym}

\usepackage[T1]{fontenc}

\usepackage[utf8]{inputenc}

\usepackage{microtype}

\usepackage{inconsolata}

\usepackage{graphicx}
\usepackage{booktabs}
\usepackage{makecell}
\usepackage[table]{xcolor}
\usepackage{comment}
\usepackage{pgfplots}
\pgfplotsset{compat=1.17}
\usepackage{amsmath}
\usepackage{caption}
\usepackage{subcaption}
\usepackage{xspace}
\usepackage{amssymb}
\usepackage{stmaryrd}
\usepackage{enumitem}
\usepackage{pifont}
\usepackage{listings}
\usepackage[ruled,vlined]{algorithm2e}

\newcommand{\et}{\textsc{ExStrucTiny}\xspace}

\title{\et: A Benchmark for Schema-Variable \\ Structured Information Extraction \\ from Document Images}

\author{
    \textbf{Mathieu Sibue}
    \quad
    \textbf{Andres Muñoz Garza}
    \quad
    \textbf{Samuel Mensah}
    \quad
    \textbf{Pranav Shetty} \\
    \textbf{Zhiqiang Ma}
    \quad
    \textbf{Xiaomo Liu} 
    \quad
    \textbf{Manuela Veloso}\\
    J.P. Morgan AI Research\\
    \texttt{\{name\}.\{surname\}@jpmchase.com}
}

\begin{document}
\maketitle

\input{latex/sections/abstract}

\input{latex/sections/intro}

\input{latex/sections/relatedwork}
\input{latex/sections/methodology}

\input{latex/sections/experiments}

\input{latex/sections/conclusion}

\input{latex/sections/limitations}

\bibliography{custom}
\input{latex/sections/appendix}

\end{document}

%% file: latex/sections/abstract.tex
\begin{abstract}

Enterprise documents, such as forms and reports, embed critical information for downstream applications like data archiving, automated workflows, and analytics. Although generalist Vision Language Models (VLMs) perform well on established document understanding benchmarks, their ability to conduct holistic, fine-grained structured extraction across diverse document types and flexible schemas is not well studied. Existing Key Entity Extraction (KEE), Relation Extraction (RE), and Visual Question Answering (VQA) datasets are limited by narrow entity ontologies, simple queries, or homogeneous document types---often overlooking the need for adaptable and structured extraction. To address these gaps, we introduce \textbf{\et},\footnote{Full dataset available upon request for research only at \url{airdata.requests@jpmorgan.com}}
a new benchmark dataset for structured Information Extraction (IE) from document images, unifying aspects of KEE, RE, and VQA. Built through a novel pipeline combining manual plus synthetic human-validated samples, \et covers more varied document types and extraction scenarios. We analyze open and closed VLMs on this benchmark, highlighting challenges such as schema adaptation, query under-specification, and answer localization. We hope this work provides a bedrock for improving generalist models for structured IE in visually rich documents.

\end{abstract}

%% file: latex/sections/intro.tex
\section{Introduction}

\input{latex/figures/line_chart_nb_ents}

\begin{figure*}[t]
    \centering
    \includegraphics[width=0.95\textwidth]{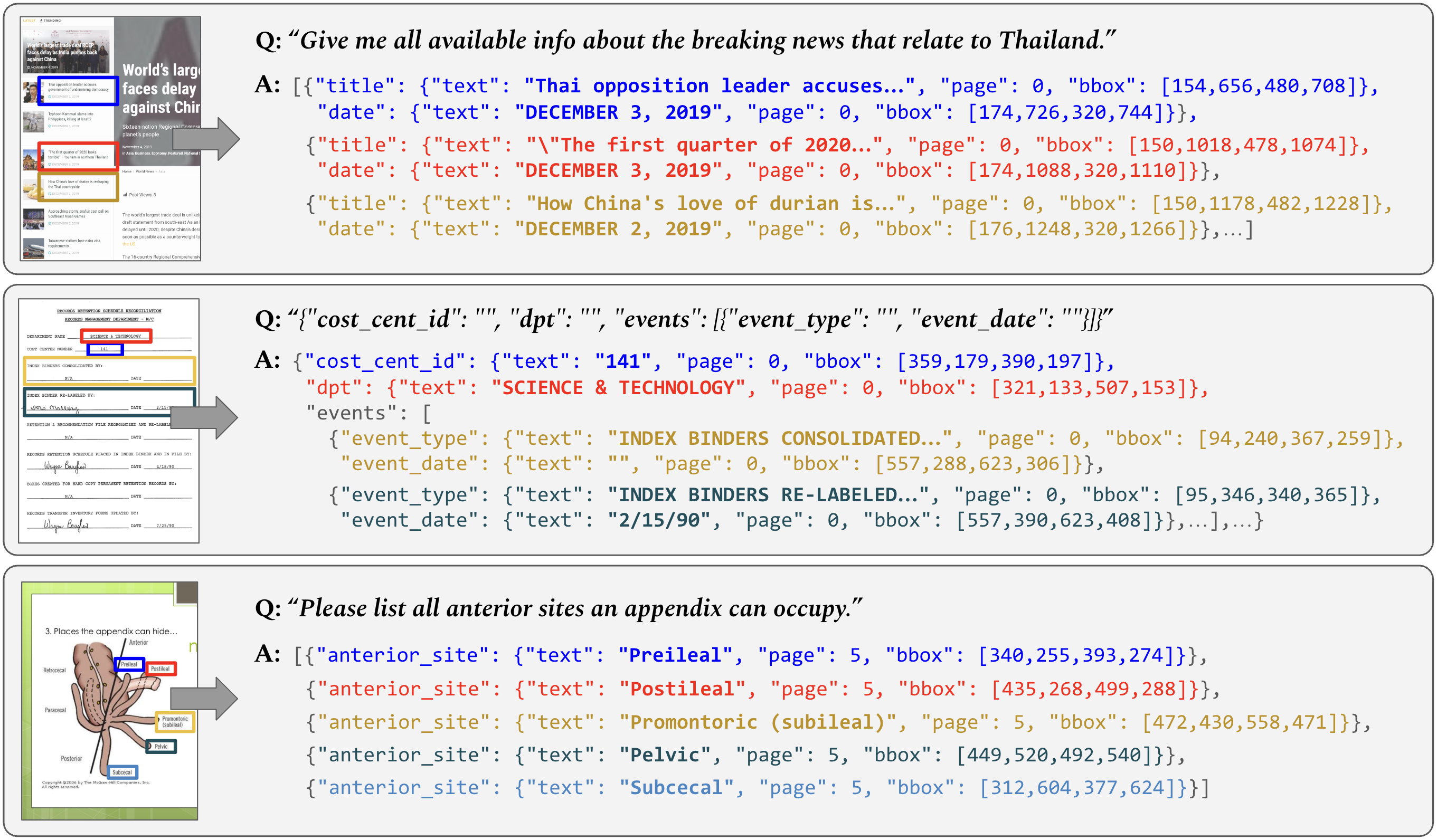}
    \label{fig:two-images}
    \caption{Three example query-answer pairs from \et. Top: on-demand IE query in plain text. Middle: closed IE query with schema. Bottom: closed IE query in plain text. Answers are JSON-formatted objects following the same structuring conventions.}
    \vspace{-0.75em}
\end{figure*}

Documents are pervasive in business operations, underpinning critical workflows in domains such as finance and healthcare. Forms, receipts, or reports routinely encode essential information that must be extracted to support data storage, automated processes, compliance, and analytics. Manual review is labor-intensive and error-prone, thus motivating research in visually rich document understanding (VRDU), at the crux of natural language processing and computer vision \cite{cui-2021-doc}.

In the text-only IE literature, extractive tasks are often categorized by the specificity of their input queries \cite{qi-etal-2024-adelie}: closed IE requires a predefined schema to be populated with relevant values, open IE seeks to discover structure without explicit guidance, and on-demand IE addresses underspecified queries that require models to extract implicitly pertinent information \cite{instructextract2023jiao}. This spectrum of query specification has shaped the design and evaluation of text-only IE models.

In the VRDU field, KEE and RE datasets are designed for closed IE, focusing on extracting a predefined set of entities from a single document type using sequence tagging approaches \cite{park2019cord, jaume-2019-funsd, wang-etal-2025-buddie}. While realistic in data-rich scenarios, these datasets for training and evaluating specialist models are not meant to assess data-scarce generalization to new domains, entities, and user requirements. On the other hand, VQA datasets in VRDU cover a broader range of entities, but their extractive questions are typically simple single-span closed IE queries \cite{mathew-2021-docvqa, Landeghem-2023-dude}, which do not measure models' ability to extract and organize multiple pieces of information in structured formats. Finally, while table and chart parsing datasets bear similarities to open IE, most solely concentrate on isolated layout elements and lack the broader document context needed for a holistic evaluation \cite{webcharts2019choi,chartqa2022masry,  pubtables2022smock}.

The emergence of LLMs and VLMs has enabled more flexible instruction-driven extraction, bridging the modeling gap between closed, open, and on-demand IE \cite{qi-etal-2024-adelie, wang-2023-instructuie, perot-etal-2024-lmdx, He-2023-ICLD3IE}. However, existing VRDU datasets do not adequately assess these models' capacity for holistic, fine-grained structured extraction across heterogeneous documents with novel entities and user-specified schemas---especially in zero and few-shot settings.

To address these limitations, we contribute (1)~\textbf{\et}, a new structured IE benchmark dataset from document images, comprising 304 closed and on-demand IE query-answer pairs and spanning 110 multi-page documents. \et unifies aspects of KEE, RE, and VQA, and is constructed through (2) a novel pipeline combining manual annotations and LLM-generated, human-validated samples across forms, financial reports, slide decks, and webpage screenshots. Additionally, we propose (3) a new evaluation framework for structured IE, and provide (4) a comprehensive performance analysis of open and closed VLMs on \et to highlight persistent challenges with schema adaptation, query underspecification, and answer localization in IE.

%% file: latex/figures/line_chart_nb_ents.tex
\begin{figure}[t]
    \begin{minipage}{0.35\textwidth}
      \centering
      \begin{tikzpicture}
        \begin{axis}[
          width=1.16\textwidth,
          height=0.9\textwidth,
          axis line style={thick},
          xlabel={\# values to extract per query},
          ylabel={ANLS},
          xlabel near ticks,
          ylabel near ticks,
          xlabel shift={-4pt},
          ylabel shift={-7pt},
          xmin=0, xmax=110,
          ymin=0, ymax=110,
          xtick={0,20,40,60,80,100},
          ytick={0,20,40,60,80,100},
          grid=both,
          grid style={gray!20},
          legend style={
            at={(0.99,0.99)},
            anchor=north east,
            font=\scriptsize,
            draw=none,
            fill=white,
            fill opacity=0.6,
            text opacity=1,
            legend columns=3
          },
          tick label style={font=\small},
          label style={font=\small},
          scale only axis
        ]
    
          \addplot+[red!75,solid,mark=o,mark size=2.2,mark options={fill=white},line width=1.2pt,error bars/.cd, y dir=both, y explicit,]
            coordinates {
              (12.5,80.3)
              (37.5,73.7)
              (62.5,85.7)
              (92.5,74.7)
            };
          \addlegendentry{closed}

          \addplot+[blue!75,solid,mark=o,mark size=2.2,mark options={fill=white},line width=1.2pt,error bars/.cd, y dir=both, y explicit]
            coordinates {
              (12.5,69.0)
              (37.5,51.9)
              (62.5,46.7)
              (92.5,29.4)
            };
          \addlegendentry{open, L}

          \addplot+[blue!60,dashed,mark=o,mark size=2.2,mark options={fill=white},line width=1.2pt,error bars/.cd, y dir=both, y explicit]
            coordinates {
              (12.5,63.7)
              (37.5,49.5)
              (62.5,61.6)
              (92.5,19.2)
            };
          \addlegendentry{open, M}

        \addlegendimage{empty legend}
        \addlegendentry{}
    
          \addplot+[blue!40,dashdotted,mark=o,mark size=2.2,mark options={fill=white},line width=1.2pt,error bars/.cd, y dir=both, y explicit]
            coordinates {
              (12.5,55.4)
              (37.5,37.3)
              (62.5,42.0)
              (92.5,9.9)
            };
          \addlegendentry{open, S}

          \addplot+[blue!25,dotted,mark=o,mark size=2.2,mark options={fill=white},line width=1.2pt,error bars/.cd, y dir=both, y explicit]
            coordinates {
              (12.5,42.7)
              (37.5,24.4)
              (62.5,22.5)
              (92.5,2.6)
            };
          \addlegendentry{open, XS}
    
        \end{axis}
      \end{tikzpicture}
    \end{minipage}

  \caption{Extraction performance declines for open source VLMs as number of values to extract increases. \vspace{-1em}}
  \label{fig:performance-comparison}
\end{figure}

%% file: latex/sections/relatedwork.tex
\section{Related Work}

\input{latex/tables/dataset_comparison}

\paragraph{IE Datasets} KEE is one of the most studied extraction task in the VRDU landscape, with numerous datasets each targeting a specific type of document, such as receipts (CORD,~\citet{park2019cord}), invoices (VRDU ad-buy,~\citet{vrdu2023wang}), and forms (FUNSD,~\citet{jaume-2019-funsd}). These datasets rely on predefined entity ontologies that are closely tied to the targeted domain, and for which values need to be extracted. For example, CORD annotates 24 specific entities in restaurant receipts (e.g., \textit{``menu item name''}).
Recent work has explored adapting KEE datasets for generative models by converting their annotations into natural language closed IE queries using either templates~\cite{zmigrod-etal-2024-value} or LLMs~\cite{boundingdocs2025giovannini}. However, most efforts in this direction produce only single-span extractive questions restricted by the fixed schema of the source datasets, making them fail to capture the complexity and diversity of real-world extractive queries. 

Document VQA datasets, such as DocVQA \cite{mathew-2021-docvqa}, InfographicVQA \cite{infovqa2022mathew}, DUDE \cite{Landeghem-2023-dude}, SlideVQA \cite{tanak-2023-slidevqa}, and VisualMRC \cite{tanak-2021-visualmrc}, often include a subset of extractive question-answer pairs akin to closed IE, but with a broader scope than KEE thanks to an ontology-agnostic design. For instance, DocVQA includes a wide variety of questions like \textit{``What is the invoice number?''} or \textit{``Who is the author of the document?''} However, these questions frequently present significant overlap with the language of the documents, which might let models answer correctly by string matching and proximity, rather than true semantic understanding.\footnote{As suggested by the high scores achieved on DocVQA, InfographicVQA or ChartQA by small VLMs \cite{qwen2.5vl2025bai}} Furthermore, VQA datasets rarely request multiple related entities in a single question and seldom include unanswerable questions \cite{Landeghem-2023-dude}, even though such cases are common in practical applications.

In contrast, \et provides closed and on-demand IE queries---including multi-entity, low-lexical-overlap, and unanswerable queries---across four document types, generated without the constraint of a fixed entity ontology. This enables robust zero-shot evaluation of VLMs for structured IE, directly addressing the limitations of previous KEE and VQA datasets. Table \ref{tab:dataset-comparison} compares \et and the test split of popular KEE and VQA datasets in VRDU across key statistics. 
\vspace{-0.2em}


\paragraph{IE Modeling} Traditionally, extractive KEE and VQA tasks in VRDU have been tackled using sequence tagging models over OCR tokens, with specialist architectures fine-tuned in data-rich settings for specific domains \cite{DLBKEFBDSLR_rombach, layoutlm2020xu}, though a few early generative exceptions existed \cite{powalski2021tilt}. These pipeline approaches, usually combining OCR, vision and text encoders, were not designed nor evaluated for zero-shot transfer to new document types, entities, and queries---thus limiting their generalization. Subsequently, vision-only autoregressive models like Donut \cite{kim-2022-donut} and Pix2Struct \cite{pix2struct2023lee} enabled flexible structured output generation thanks to the use of a decoder, but lacked robust instruction-following capabilities for fulfilling diverse user intents. The more recent post-trained LLMs \cite{perot-etal-2024-lmdx, He-2023-ICLD3IE} and VLMs \cite{hu-2024-mplug-docowl, wang-etal-2024-docllm, bai-2023-qwenvl, pixtral} shift the field toward zero-shot structured IE, where models can adapt to novel entities or schemas specified in user queries on-the-fly. However, to our knowledge, no existing VRDU benchmark supports evaluating such generalist models for generic extraction. \et directly addresses this by enabling comprehensive few-shot evaluation of VLMs for structured IE over document images.

\vspace{-1em}
\paragraph{Synthetic Data Generation for IE}
Recent studies have highlighted that LLMs can efficiently generate high-quality 
instruction following data, when properly prompted \citep{HonovichSLS23, itgpt42023peng, alpaca2023, WangKMLSKH23}. While similar approaches have been successfully applied in text-only IE \citep{qi-etal-2024-adelie, nuextract, relprompt2022chia}, their use in VRDU is still emerging, as seen in works like \citet{boundingdocs2025giovannini}, which uses LLMs to create extractive questions from KEE annotations. However, as noted earlier, these methods inherit limitations from source annotations. Despite this, LLM-based data generation remains a promising avenue for scaling and enriching benchmarks \citep{KimSY0LWGLWN25}, and we leverage this strategy to build \et.
\vspace{-0.3em}

%% file: latex/tables/dataset_comparison.tex
\begin{table*}[h!]
\centering\resizebox{\textwidth}{!}{
\begin{tabular}{lrrrrrrrrrr}
\toprule
\bf Dataset &\bf  \#Docs &\bf  \#QAs &\bf  Len(Doc) &\bf Len(Q) &\bf  Len(A) &\bf  \#Entities per Q &\bf  \#Values per A &\bf  \%Q Unansw. &\bf  Q-Doc Overlap &\bf  $\overline{|\text{Ontology}|}$ \\ \midrule
CORD & 100 & 100 (2400) & 23.6 & 64.0 (2.7) & 24.0 (1.0) & 24.0 (1.0) & 13.4 (0.6) & 100\% (67\%) & 39\% (46\%) & 0.24 (0.01) \\ 
Kleister Charity & 606 & 606 (4848) & 6824.6 & 24.0 (3.0) & 9.2 (1.2) & 8.0 (1.0) & 4.6 (0.6) & 100\% (43\%) & 48\% (69\%) & 0.01 (0.00) \\ 
VRDU Ad-buy & 641 & 641 (8974) & 1524.4  & 25.0 (1.8) & 103.5 (7.4) & 14.0 (1.0) & 65.1 (4.7) & 95\% (16\%) & 44\% (75\%) & 0.02 (0.00) \\ 
VRDU Registration & 1915  & 1915 (11490) & 948.8  & 13.0 (2.2) & 13.4 (2.2) & 6.0 (1.0) & 4.5 (0.7) & 83\% (26\%) & 55\% (74\%) & 0.00 (0.00) \\
FUNSD & 47 & 47 (32430) & 253.9  & 1652.0 (2.4) & 80.3 (0.1) & 690.0 (1.0) & 13.0 (0.1) & 100\% (98\%)& 42\% (59\%) & 14.7 (0.02) \\  \midrule
DocVQA & 1287 & 4733 & 194.9 & 8.5 & 2.1 & 1.0 & 1.0 & 0\% & 56\% & 0.85 \\ 
SlideVQA$_{sample}$ & 218 & 760 & 1252.0 & 12.6 & 1.7 & 1.1 & 1.0 & 0\% & 62\% & 1.00\\
DUDE$_{val}$ & 732 & 1986 & 1745.7 & 8.4 & 3.3 & 1.1 & 1.2 & 13\% & 61\% & 1.00  \\
TAT-DQA & 277 & 662 & 544.6 & 10.7 & 7.8 & 1.0 & 1.0 & 0\% &67\% & 1.06 \\
VisualMRC & 1392 & 1472 & 1017.5 & 8.8 & 3.8 & 1.0 & 1.0 & 0\% &68\% & 1.02 \\
\midrule
{\et} & 110 & 304  & 1278.5  & 18.2 & 33.5  & 4.0  & 8.7  & 41\% & 47\%  & 3.46 \\ 
\bottomrule
\end{tabular}}

\caption{Comparison of \et VS. test splits of KEE (top) and VQA (bottom) datasets across key statistics: number of documents (\textit{\#Docs}), number of extractive QA pairs (\textit{\#QAs}), average document length (\textit{Len(Doc)}), average query length (\textit{Len(Q)}), average answer length (\textit{Len(A)}), average number of entities requested per query (\textit{\#Entities per Q}), average number of extracted values in the answer (\textit{\#Values per A}), percentage of queries requesting some missing entities (\textit{\%Q Unansw.}), lexical overlap (\textit{Q-Doc Overlap}), normalized ontology size (\textit{$\overline{|\text{Ontology}|}$}). As the test split of DUDE does not include answers, we use the validation split instead. See Appendix \ref{appendix:dataset-comparison} for more details.}
\label{tab:dataset-comparison}
\vspace{-1em}
\end{table*}

%% file: latex/sections/methodology.tex
\section{Methodology}

\input{latex/sections/task_formulation}

\input{latex/sections/manual_annotation}

\input{latex/sections/data_generation}

\input{latex/sections/data_validation_2}

\input{latex/sections/schema_mapping}


%% file: latex/sections/task_formulation.tex
\vspace{-0.3em}
\subsection{Task Formulation}

We mathematically formalize the structured extraction task in \et to highlight its distinction from traditional KEE, RE, and VQA in VRDU. Unlike prior benchmarks, \et requires models to extract and organize multiple pieces of information according to variable user-specified schemas, rather than a fixed ontology of entities. Each query-answer pair $(\mathbf{x}_i, \mathbf{y}_i)$ consists of:
\vspace{-0.4em}
\begin{itemize}[leftmargin=1em]
    \setlength\itemsep{-0.2em}
    \item An input set of $d$ document images $\{I_{i,j}\}_{j=1}^d$;
    \item An input prompt $\mathbf{q}_i$ (string), specifying the entities to extract, either in the form of descriptive text or of a schema to populate;
    \item A ground truth structured output $\mathbf{y}_i$ (string), which includes the verbatim text values $\{t_{i,k}\}_{k=1}^n$ extracted for the requested entities, their page indices $\{p_{i,k}\}_{k=1}^n$ ($p_{i,k}\! \in\! \llbracket 1, d \rrbracket$), and bounding boxes $\{b_{i,k}\}_{k=1}^n$ ($b_{i,k}\! \in\! \llbracket 0, 1000 \rrbracket^4$).
\end{itemize}
\setlength{\abovedisplayshortskip}{2pt}
The structured output can be expressed as follows:
$$
\mathbf{y}_i = s_i\left(\{t_{i,k}\}_{k=1}^n, \{p_{i,k}\}_{k=1}^n, \{b_{i,k}\}_{k=1}^n\right)
$$
where $s_i$ is a function that populates the extracted values, page indices, and bounding boxes into a schema fulfilling the query $\mathbf{q}_i$. We call each $(t_{i,k}, p_{i,k}, b_{i,k})$ an \textit{extraction leaf} of answer $\mathbf{y}_i$.

Given a vision-language model $f_\theta$, the objective is to generate a string $\hat{\mathbf{y}}_i$ that can be parsed into a structured object containing extractions for the entities specified in $\mathbf{q}_i$:
$$
\hat{\mathbf{y}}_i = f_\theta\left(\{I_{i,j}\}_{j=1}^d, \mathbf{q}_i \right)
$$
with
$
\hat{\mathbf{y}}_i = \hat{s}_i(\{\hat{t}_{i,k}\}_{k=1}^m, \{\hat{p}_{i,k}\}_{k=1}^m, \{\hat{b}_{i,k}\}_{k=1}^m)
$.

Section~\ref{sec:schema_mapping} introduces metrics to evaluate model performance on structure prediction ($s_i$), text extraction ($\{t_{i,k}\}_{k=1}^n$), page localization ($\{p_{i,k}\}_{k=1}^n$), and bounding box prediction ($\{b_{i,k}\}_{k=1}^n$).

%% file: latex/sections/manual_annotation.tex
\subsection{Small-Scale Manual Data Annotation}
\label{sec:manual-data-annot}

\input{latex/figures/entities_wordcloud}

To construct an initial subset of \et, we manually annotate query-answer (QA) pairs supporting both closed and on-demand IE---which cover an adequate range of query specificity.
More precisely, we include three query types:\footnote{We use ``query type'' and ``QA type'' interchangeably.}
\begin{itemize}[leftmargin=1em]
    \setlength\itemsep{-0.2em}
    \item \textbf{Closed with plain text:} explicitly requests in plain text the values of specific entities (e.g., \textit{``Extract the name and role of the signers.''})
    \item \textbf{Closed with schema:} provides a JSON-formatted string schema with empty leaves to populate with the values of specific entities (e.g., \textit{`[\{"signer name": "", "signer role": ""\}]'})
    \item \textbf{On-demand with plain text:} implicitly asks for the values of child attributes (not known in advance) of parent entities (e.g., \textit{``Extract all details about the signers.''}).
\end{itemize}

\begin{figure*}[t!]
    \centering
    \includegraphics[width=0.9\textwidth]{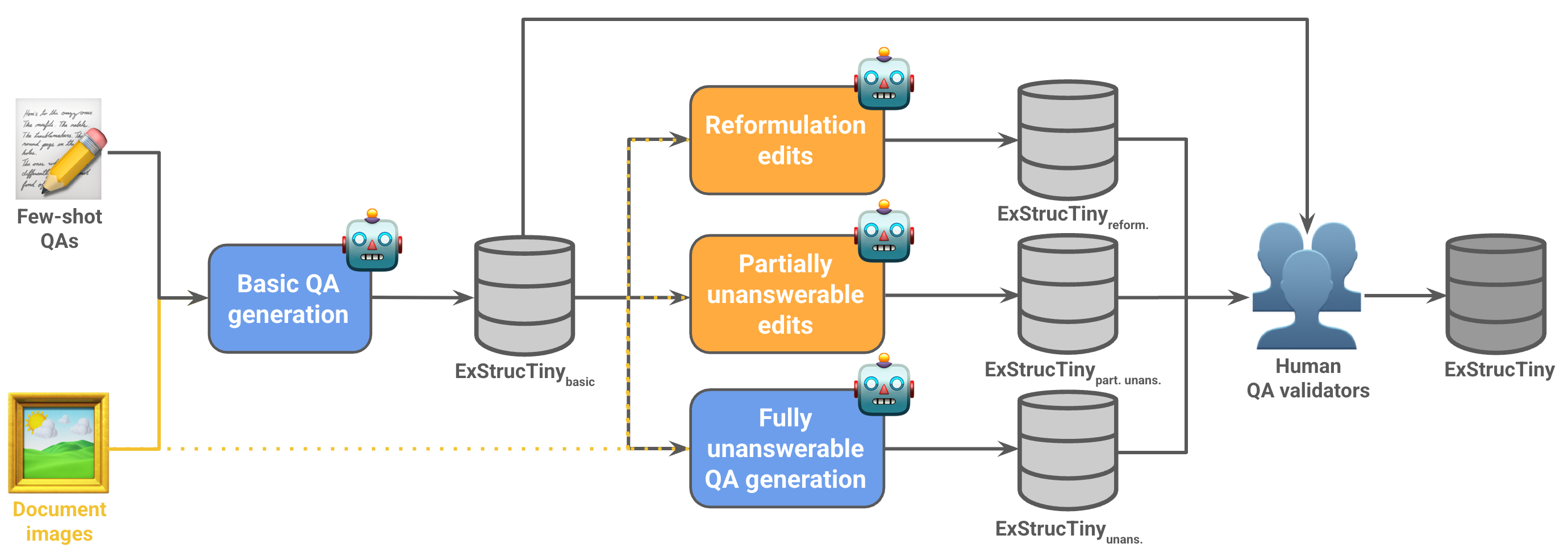} 
    \caption{\et Synthetic Data Generation Pipeline}
    \label{fig:synth-data-gen}
    \vspace{-0.75em}
\end{figure*}

We require \textbf{all answers to be in JSON format}, regardless of the query type. This enforces structured extraction, enables the representation of relationships between entities, and aligns with common industry practices for API and pipeline communication (where JSON format is prevalent). To minimize ambiguity in output structuring, we provide detailed annotation instructions (Appendix \ref{app:data-gen-instructions}) ensuring each extraction leaf $(t_{i,k}, p_{i,k}, b_{i,k})$ is represented as a leaf object \texttt{\{"text":}~$t_{i,k}$\texttt{,"page":}~$p_{i,k}$\texttt{,"bbox":}~$b_{i,k}$\texttt{\}} in the JSON output. If an entity is not present in the document, its text, page, and bbox fields are set to \texttt{null}; if the value is empty but the entity is present (e.g., a blank form field), only the text value is \texttt{""}. For checkboxes, values are standardized as \texttt{"Yes"} or \texttt{"No"}. Example QA pairs are shown in Figure~2.\footnote{A closed-with-plain-text counterpart of the middle example could be \textit{``What are the cost cent id, dpt, and events (each with a type and date)?''}}

To maximize variety in layout, length, and content, we source documents from the test splits of four existing datasets: FUNSD (\citet{jaume-2019-funsd}, forms), TAT-DQA (\citet{tatdqa2022zhu}, financial reports), SlideVQA (\citet{tanak-2023-slidevqa}, long slide decks), and VisualMRC (\citet{tanak-2021-visualmrc}, webpage screenshots). Four annotators each select three random documents from each dataset and generate $\sim\!3$ QAs of each type per document, following strict guidelines to ensure realism and challenge: queries should (1) request multiple related entities,  (2) include some entities likely to be missing---reflecting real-world scenarios---, (3) exhibit limited lexical overlap with document text, (4) target entities with values spanning multiple pages when possible, and (5) incorporate less common layout elements, such as charts and checkboxes. Annotators are also advised not to reuse annotations of the source datasets to avoid repetition and potential test leakage.

Created QAs are then reviewed by a separate annotator to ensure correct extractions as well as consistency with the output formatting and structuring instructions. This process resulted in 102 high-quality QA pairs, spanning a satisfactory range of layouts and extraction challenges.

%% file: latex/figures/entities_wordcloud.tex

\begin{figure}[t!]
    \centering
    \resizebox{0.9\columnwidth}{!}{%
        \includegraphics[width=\textwidth]{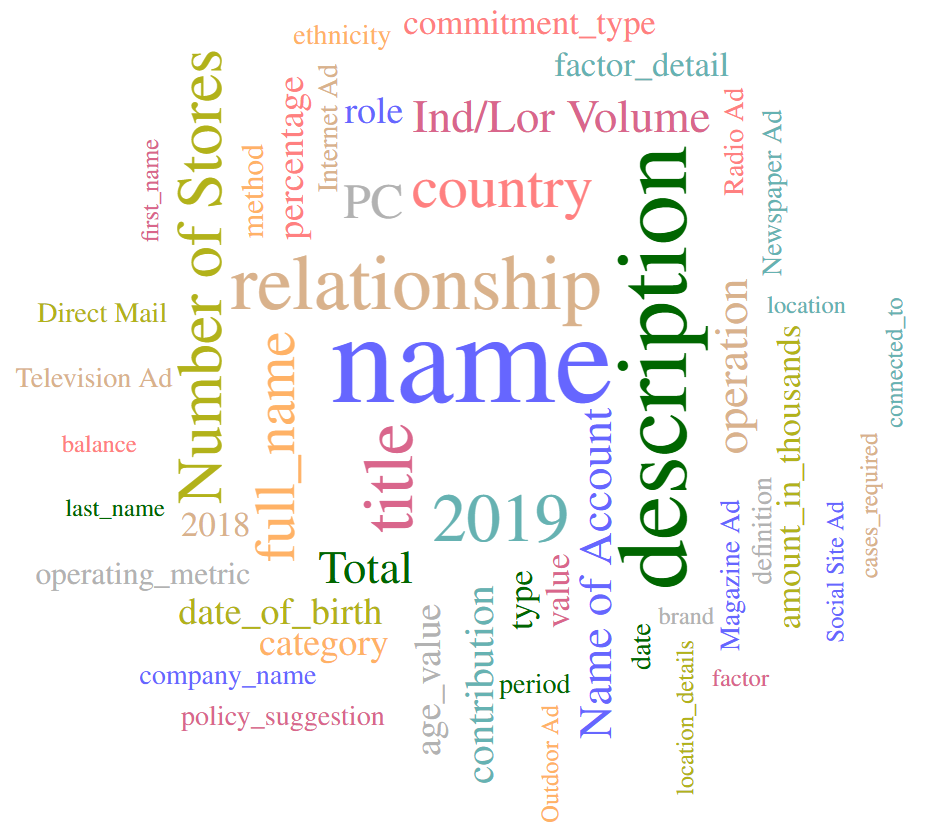}
    }
    \vspace{-0.75em}
    \caption{Example entities requested in \et}
    \vspace{-1em}
    \label{fig:word_cloud}
\end{figure}

%% file: latex/sections/data_generation.tex
\subsection{Large-Scale Synthetic Data Generation}

Manual annotation, while high-quality, does not scale efficiently for cheaply generating diverse and difficult QAs. To address this, we leverage \texttt{Gemini-2.5-Flash-Thinking} \cite{gemini2025}, a competitive large VLM, to synthetically generate QAs and expand our benchmark.

Specifically, for each source dataset, we provide \texttt{Gemini} with thorough instructions of our task as well as few-shot examples from the manually-annotated QAs for this dataset (Section \ref{sec:manual-data-annot}). The model is then iteratively prompted with new document images to generate QAs of each type, while avoiding duplication. We empirically observe that accompanying each shot with a chain-of-thought (CoT) explanation \cite{cot2022wei}---to detail the process behind the creation of each QA given its document---improves adherence to our output structuring guidelines. We also find that using a higher temperature (0.8) and a large reasoning budget (4096 tokens) helps strike a good balance between diversity and quality in the QAs generated.

As illustrated in Figure~\ref{fig:synth-data-gen}, the initial set of synthetic QAs is subsequently enriched by optionally sending it through different augmentation streams, where the VLM is reprompted to generate new and modified QAs with focused characteristics. On the one hand, we instruct \texttt{Gemini} to reformulate entity names in queries and their corresponding answers, thereby reducing lexical overlap with document text. On the other hand, we ask \texttt{Gemini} to introduce new entities that are absent from the source document, simulating realistic scenarios where requested information is missing. We also prompt \texttt{Gemini} to generate from scratch fully unanswerable QAs, where all requested entities are missing. These enrichments directly support our dataset desiderata by ensuring coverage of many real-world extraction cases.

Finally, to achieve a balanced and representative dataset, we subsample synthetic QAs from each pipeline stage, targeting proportions of 55\% basic, 25\% reformulated, 15\% partially unanswerable, and 5\% fully unanswerable QAs. These proportions are configurable hyperparameters of the QA generation pipeline and can be rebalanced to match specific testing objectives in the future.

%% file: latex/sections/data_validation_2.tex
\input{latex/tables/changes_table}

\subsection{Dataset Quality Validation}
\label{sec:data_val}

To guarantee the correctness and consistency of each synthetic QA instance, we conduct a manual validation exercise using a custom-built UI. Three expert validators with prior IE and VQA experience review up to 100 QAs each, focusing on five key areas: clarifying and grounding query text, ensuring extracted text matches the document verbatim, correcting page references, refining bounding boxes, and enforcing output structuring rules. Synthetic QAs were prioritized for review based on entity rarity, layout complexity, and number of entities (see Algorithm~\ref{alg:target_guided_sampling_simple} in Appendix~\ref{apx:dataset-annotation-details}).

Rather than rejecting borderline cases, validators were tasked with repairing errors in query text, extracted text, page references, bounding boxes, and answer schema whenever possible. QAs were discarded only when errors were irreparable, such as ungrounded questions or non-trivial malformed JSONs; in practice, only 2 samples were rejected.

After validation, we analyze the types and frequency of modifications done by validators. Overall, our budget allowed to validate 202 synthetic QAs kept in \et, with each requiring on average 25.5 edits. Table~\ref{table:modifications_summary} categorizes edits into three main classes---additions, deletions, and replacements---with 11 sub-classes. Value changes (regarding the text, page, and bbox fields of the extraction leaves) and additions account for respectively 47.8\% and 23\% of all modifications, underscoring the importance of human intervention.

To further verify dataset quality, we perform cross-validation of edits among validators; we observe a high agreement rate of 96.2\% at the extraction leaf level, indicating strong instruction adherence and robust dataset quality (more in App.~\ref{apx:dataset-annotation-details}).

After combining manually-annotated and VLM-generated human-validated instances, \et totals 304 QAs over 110 documents, with an acceptably balanced proportion of closed with plain text, closed with schema, and on-demand QAs (Table~\ref{table:final-statistics}). Figure~\ref{fig:word_cloud} highlights the top 50 entities included in it and illustrates the variety of topics covered.
Finally, Table~\ref{tab:dataset-comparison} shows that, despite fewer documents than many KEE and VQA datasets in VRDU, \et exhibits lower query-document lexical overlap, more entities requested per query, and a reasonable rate of partial unanswerability.
These characteristics pose unique challenges to VLMs for structured IE, which we analyze in detail in Section \ref{sec:expts} with benchmark results covering multiple models.

%% file: latex/tables/changes_table.tex
\begin{table}[th]
\centering
\scriptsize
\setlength\tabcolsep{3pt}
\resizebox{0.9\columnwidth}{!}{
\begin{tabular}{l c c}
    \toprule
    \textbf{Modification Type} & \textbf{Count} & \textbf{Proportion} \\
    \midrule
    \texttt{ADD} &  &  \\
    \quad\texttt{value\_added} & 1186 & 23.0\% \\
    \quad\texttt{entity\_added} & 294 & 5.7\% \\
    \quad\texttt{dict\_added} & 242 & 4.7\% \\
    \quad\texttt{list\_added} & 183 & 3.6\% \\
    \cmidrule(lr){1-3}
     Sub-total & 1905 & 37.0\% \\
    \midrule
    \texttt{DELETE} &  &  \\
    \quad\texttt{value\_deleted} & 400 & 7.8\% \\
    \quad\texttt{entity\_deleted} & 87 & 1.7\% \\
    \quad\texttt{dict\_deleted} & 79 & 1.5\% \\
    \quad\texttt{list\_deleted} & 58 & 1.1\% \\
    \cmidrule(lr){1-3}
    Sub-total & 624 & 12.1\% \\
    \midrule
    \texttt{REPLACE} &  &  \\
    \quad\texttt{value\_change} & 2465 & 47.8\% \\
    \quad\texttt{query\_type\_change} & 115 & 2.2\% \\
    \quad\texttt{entity\_renamed} & 44 & 0.9\% \\
    \cmidrule(lr){1-3}
    Sub-total & 2624 & 50.9\% \\
    \midrule
    \textbf{Total Modifications} & \textbf{5153} & \textbf{100.0\%} \\
    \bottomrule
\end{tabular}
}
\caption{Hierarchical breakdown of edits done during the data validation exercise, by type and category.}
\label{table:modifications_summary}
\vspace{-1em}
\end{table}

%% file: latex/sections/schema_mapping.tex
\input{latex/tables/final_data_stats}

\subsection{Evaluation Metrics for Structured IE}
\label{sec:schema_mapping}

To evaluate model predictions on \et, we assess performance across four dimensions for each query-answer pair $(\mathbf{x}_i, \mathbf{y}_i)$: structure prediction ($s_i$), text extraction ($\{t_{i,k}\}_{k=1}^n$), page localization ($\{p_{i,k}\}_{k=1}^n$), and bounding box prediction ($\{b_{i,k}\}_{k=1}^n$). The main challenge is that, despite detailed output structuring guidelines, VLMs may produce valid yet differently structured outputs, with possible variations in key names, extracted values, and schema organization. Indeed, multiple JSON structures may convey the same information, so exact string matching or JSON object equality is insufficient for a fair evaluation.

To address this, we propose a semantic mapping approach to align the extraction leaves $\left(\{t_{i,k}\}_{k=1}^n, \{p_{i,k}\}_{k=1}^n, \{b_{i,k}\}_{k=1}^n\right)$ from the ground truth with those from the predicted output $(\{\hat{t}_{i,k}\}_{k=1}^m, \{\hat{p}_{i,k}\}_{k=1}^m, \{\hat{b}_{i,k}\}_{k=1}^m)$. Specifically, we flatten both the ground truth and predicted structured objects---representing each leaf node by its path from the root (full entity name). These flattened objects are then input to a text-only LLM, which is few-shot prompted to generate a mapping from ground truth keys to predicted keys, accounting for paraphrasing and structural differences. Unmatched ground truth keys are mapped to \texttt{null}, indicating missing extractions.

We benchmark three small text-only LLMs \cite{llama3, phi4reas, gptoss} for this mapping task essential to the metric computation, finding that reasoning models yield the best precision and recall (Table~\ref{table:map_metrics}). More precisely, \texttt{gpt-oss-20b} provides near perfect performance for a reasonable runtime. More details on this comparison of text-only LLMs for mapping ground truth to predicted schemas are in Appendix \ref{app:eval-metric}. We use \texttt{gpt-oss-20b} with a temperature of 0.95, top-K=10, and medium reasoning effort as our schema mapper in the rest of the paper. We recommend this setup to readers for consistency.

\input{latex/tables/llm_judge_comparison}
\input{latex/tables/anls_main}

Once the mapping is established by the text-only LLM, we compute standard metrics to measure how well the predicted extractions cover the ground truth ones: recall, precision, and F1 for matched entities. Then, for each matched extraction leaf, we compute ANLS \cite{stvqa} for text similarity; page accuracy for localization; Intersection over Union (IoU, \citet{iou2005}) and normalized proximity \cite{nourbakhsh-etal-2025} for bounding box overlap. If a ground truth extraction leaf is not matched, we assign zero scores for ANLS, page accuracy, IoU, and proximity. Additionally, we use normalized tree-edit distance \cite{zss1989} to compare the overall structure of predicted and ground truth objects.

This evaluation framework enables fair and comprehensive assessment of VLMs on \et---accounting for the inherent diversity in structured output generation.

%% file: latex/tables/final_data_stats.tex
\begin{table}[t]
\centering\resizebox{0.9\columnwidth}{!}{
\begin{tabular}{lcc}
\toprule
\textbf{Statistic} & \textbf{Manual} & \textbf{Synthetic} \\
\midrule
\# documents & 12 & 98 \\
Avg. \# QAs / doc & 8.5 & 2.1 \\
Avg. annot. + valid. time & 50 min & 16 min \\
\# QAs & 102 & 202 \\
\quad\quad Closed w/ plain text & 51 & 68 \\
\quad\quad Closed w/ schema & 25 & 68 \\
\quad\quad On-demand & 26 & 66 \\
\bottomrule
\end{tabular}}
\caption{Final manual vs. synthetic data comparison.}
\label{table:final-statistics}
\end{table}

%% file: latex/tables/llm_judge_comparison.tex
\begin{table}[t]
\centering
\scriptsize
\setlength\tabcolsep{3pt}
\resizebox{1.0\columnwidth}{!}{
\begin{tabular}{l c c c c c}
    \toprule
    \textbf{Model} & \textbf{Reason.} 
    & \makecell{\textbf{Map.}\\\textbf{R}} 
    & \makecell{\textbf{Map.}\\\textbf{P}} 
    & \makecell{\textbf{Map.}\\\textbf{F1}}
    & \makecell{\textbf{Total}\\\textbf{ time (s)}} \\
    \midrule
    \texttt{Llama-3.1-8b} &  \ding{55}   & 0.960 & 0.784 & 0.863 & \textbf{560} \\
    \texttt{Phi-4-reasoning-14B} & \ding{51}   & \underline{0.967} & \underline{0.924} & \underline{0.945} & 7741\\
    \texttt{gpt-oss-20b} &   \ding{51}  & \textbf{0.996} & \textbf{0.956} & \textbf{0.976} &  \underline{1552} \\
    \bottomrule
\end{tabular}
}
\caption{Recall, precision, and F1 for schema mappers.}
\label{table:map_metrics}
\vspace{-1em}
\end{table}

%% file: latex/tables/anls_main.tex
\begin{table*}[ht!]
    \centering
    \small
    \setlength\tabcolsep{4pt}
    \resizebox{\textwidth}{!}{
    \begin{tabular}{l ccc c ccc c ccc c ccc c c c c}
        \toprule
        & \multicolumn{3}{c}{\textbf{Forms}} & & 
        \multicolumn{3}{c}{\textbf{Financial Reports}} & & 
        \multicolumn{3}{c}{\textbf{Slide Decks}} & & 
        \multicolumn{3}{c}{\textbf{Web Pages}} & & 
        & & \\ 
        \cmidrule(lr){2-4} \cmidrule(lr){6-8} \cmidrule(lr){10-12} \cmidrule(lr){14-16}
        \textbf{Model} 
        & text & schema & on-dem. & & 
        text & schema & on-dem. & & 
        text & schema & on-dem. & & 
        text & schema & on-dem. & & 
        \textbf{Manual} & \textbf{Synthetic} & \textbf{Avg.} \\ 
        \midrule
        \texttt{Qwen2.5-VL-3B}          
        & 61.4 & 48.3 & 47.1 & & 
        50.4 & 56.2 & 21.5 & & 
        55.5 & 40.4 & 32.4 & & 
        46.5 & 21.9 & 24.2 & & 
        35.0 & 40.1 & 38.6 \\
        \texttt{Gemma-3-4B}             
        & 64.8 & 48.3 & 25.3 & & 
        37.3 & 46.6 & 12.6 & & 
        33.9 & 27.3 & 9.9 & & 
        25.8 & 16.6 & 6.0 & & 
        24.3 & 25.3 & 25.0 \\
        \midrule
        \texttt{Qwen2.5-VL-7B}          
        & 74.7 & 60.4 & 58.0 & & 
        68.3 & 63.1 & 32.4 & & 
        64.0 & 61.5 & 33.1 & & 
        45.6 & 35.9 & 26.8 & & 
        45.3 & 50.4 & 48.8 \\
        \texttt{Gemma-3-12B}            
        & 69.3 & 68.0 & 66.2 & & 
        70.5 & 59.8 & 36.4 & & 
        69.7 & 49.3 & 24.2 & & 
        46.4 & 21.2 & 19.5 & & 
        40.7 & 45.8 & 44.3 \\
        \texttt{Pixtral-12B}            
        & 70.5 & 60.6 & 74.9 & & 
        70.6 & 47.1 & 40.4 & & 
        68.0 & 53.1 & \underline{42.9} & & 
        8.9  & 4.8  & 5.4 & & 
        37.0 & 44.5 & 42.3 \\
        \texttt{Kimi-VL-A3B-16B}        
        & 67.3 & 64.5 & 67.9 & & 
        77.6 & 51.8 & 29.5 & & 
        49.3 & 42.1 & 40.3 & & 
        41.1 & 10.5 & 15.5 & & 
        32.4 & 44.8 & 41.0 \\
        \midrule
        {\color{gray}\texttt{Qwen2.5-VL-32B}$_{text}$}  & {\color{gray}73.0} & {\color{gray}57.0} & {\color{gray}39.0} & & 
        {\color{gray}78.1} & {\color{gray}72.9} & {\color{gray}\textbf{57.3}} & & 
        {\color{gray}55.1} & {\color{gray}50.0} & {\color{gray}28.7} & & 
        {\color{gray}\underline{61.3}} & {\color{gray}\underline{50.7}} & {\color{gray}29.9} & & 
        {\color{gray}48.1} & {\color{gray}50.7} & {\color{gray}49.9} \\
        \texttt{Qwen2.5-VL-32B}         
        & 67.0 & \textbf{79.3} & 66.8 & & 
        67.1 & 66.4 & 39.6 & & 
        45.0 & \underline{69.8} & 37.4 & & 
        49.3 & 48.3 & \underline{38.5} & & 
        40.6 & \underline{61.3} & 55.1 \\
        \texttt{Gemma-3-27B}            
        & \textbf{75.7} & 70.5 & 70.6 & & 
        70.6 & 61.8 & 52.7 & & 
        \textbf{83.7} & 64.3 & 36.5 & & 
        50.0 & 24.6 & 21.6 & & 
        \underline{50.0} & 54.1 & 52.9 \\
        \texttt{Mistral-Small-3.2-24B}  
        & \underline{75.0} & \underline{78.5} & \underline{80.2} & & 
        \textbf{88.9} & \textbf{78.5} & 41.5 & & 
        \underline{83.5} & \textbf{85.7} & \textbf{47.2} & & 
        18.9 & 12.7 & 7.4 & & 
        45.0 & 60.3 & \underline{55.8} \\
        \midrule
        \texttt{Qwen2.5-VL-72B-FP8}         & 74.5 & 70.9 & \textbf{80.3} & & 
        \underline{83.0} & \underline{70.2} & \underline{56.9} & & 
        78.7 & 63.5 & 42.4 & & 
        \textbf{67.1} & \textbf{53.2} & \textbf{39.6} & & 
        \textbf{58.8} & \textbf{62.5} & \textbf{61.4} \\
        \midrule
        \midrule
        \rowcolor{gray!20}
        \texttt{Gemini-2.5-Flash}       
        & \underline{82.1} & \textbf{86.9} & \textbf{85.9} & & 
        \underline{88.7} & \underline{80.9} & \underline{55.8} & & 
        \underline{78.2} & \textbf{86.2} & \underline{71.5} & & 
        \underline{68.1} & \underline{65.7} & \underline{43.6} & & 
        \underline{67.8} & \underline{76.1} & \underline{73.6} \\
        \rowcolor{gray!20}
        \texttt{Gemini-2.5-Pro}         
        & \textbf{85.3} & \underline{83.2} & \underline{75.0} & & 
        \textbf{91.0} & \textbf{82.5} & \textbf{80.0} & & 
        \textbf{91.5} & \underline{84.4} & \textbf{80.4} & & 
        \textbf{74.7} & \textbf{72.0} & \textbf{63.0} & & 
        \textbf{81.2} & \textbf{78.8} & \textbf{79.5} \\
        \bottomrule
    \end{tabular}
    }
    \caption{Model performance on text extraction (ANLS) across document types and query types.}
    \label{table:results_grouped}
    \vspace{-1em}
\end{table*}

%% file: latex/sections/experiments.tex
\section{Experiments}
\label{sec:expts}

\input{latex/sections/experiment_settings}

\input{latex/sections/extraction_results}

\input{latex/sections/location_experiment}
\input{latex/tables/localization}

\input{latex/sections/structure_experiments}
\input{latex/tables/structure}

\input{latex/sections/answer_context_experiment}

%% file: latex/sections/experiment_settings.tex
\subsection{Settings}


We benchmark closed \cite{gemini2025} and open VLMs \cite{qwen2.5vl2025bai, gemma3, pixtral, mistralsmall, kimivl} of various sizes and architectures from multiple providers on \et. We use vLLM \cite{vllm2023} for inference on four NVIDIA L40S GPUs with a temperature of 0.2 for all open models. Evaluation uses metrics from Section \ref{sec:schema_mapping}, ensuring consistent, thorough comparison. Each VLM receives three shots from \et (one per supported query type) to illustrate our detailed task instructions.

%% file: latex/sections/extraction_results.tex
\subsection{Text Extraction Results}
\paragraph{Closed Models Have an Edge} Results in Table~\ref{table:results_grouped} show a substantial performance gap between closed-source and open-source models across all document classes. The highest performing closed model outscores the best open model by 18+ points. 

\paragraph{Open Models Highlight the Benefit of Size} We observe a clear positive correlation between performance and model size across all benchmarked models. This relationship is particularly evident within individual model families. For example, the \texttt{Qwen2.5-VL} family shows an increase in performance from 38.5 points with the 3B parameter model to 61.4 points with the 72B parameter model. Similarly, the \texttt{Gemma-3} family starts at 25 points with the 3B version and reaches 52.9 points with the 27B parameter version.

\paragraph{On-demand and Schema Struggles}
Table~\ref{table:results_grouped} reveals that, across model sizes, performance on closed-with-schema and on-demand QAs is consistently lower than on plain-text closed QAs. The difficulty of schema queries can be attributed to their length---they request approximately three times as many entities as plain-text queries. On-demand queries are even more challenging due to their underspecification, requiring models to interpret the query in-depth, resolve implicit references to entities or concepts, and align these with the image to produce the appropriate extractions. Consequently, on-demand queries are the most difficult overall, with the majority of models recording their lowest scores on these tasks across document types.

\paragraph{Open Models are Affected by Extraction Length}
Analysis of Figure~\ref{fig:performance-comparison} shows closed-source models maintain stable extraction quality even when the QAs expect 50+ extracted values, while open-source models' performance declines as extraction count increases.
It is difficult to attribute this advantage to any specific factor, since little is known about closed models' training data, size, or training recipe. The improved robustness may result from progress across any or all of these dimensions (e.g., Table~\ref{table:results_grouped} shows  model size alone can have substantial impact within model families).

\paragraph{Manual vs. Synthetic Data}
Table~\ref{table:results_grouped} also indicates a modest performance gap between manually generated data and synthetic data. Models score on average 13.6\% lower on manually-authored QAs, suggesting they remain slightly more difficult than synthetic QAs. Crucially, there is no evidence of bias favoring \texttt{Gemini-2.5-Flash}, despite its role in drafting the synthetic data. This may be explained by the rigorous manual editing and validation performed on all \texttt{Gemini}-generated entries and detailed in Section~\ref{sec:data_val}. Finally, neither split is saturated: the best model reaches 81.2\% ANLS on the manual subset and 78.8\% on the synthetic one.

\input{latex/figures/bar-chart-synthetic-qas}

\paragraph{Reformulations and Unanswerable Requests in Synthetic Data}
Figure~\ref{fig:synthetic-breakdown} breaks down extraction performance by synthetic category: basic queries; fully and partially unanswerable ones; reformulated ones (see pipeline in Figure~\ref{fig:synth-data-gen}). VLMs seem to struggle to correctly identify when a requested entity is missing, as shown by a lower score on fully and partially unanswerable synthetic queries compared to basic ones. Similarly, ANLS drops by nearly 27\% on reformulated queries relative to basic ones, highlighting sensitivity to synonymy and reduced lexical overlap with source documents. These results underscore the value of augmenting initial \texttt{Gemini} generations through multi-stage edits to increase both the realism and the difficulty of \et.

\paragraph{Text-only Baseline} To quantify the contribution of visual information in \et, we construct a text-only baseline by applying OCR to each document and evaluating \texttt{Qwen2.5-VL-32B} using only the text tokens extracted. Compared with the image-only configuration, the text-only variant yields a $\sim$10\% lower ANLS overall---indicating visual cues materially improve structured IE performance in this benchmark (especially for on-demand QAs).

%% file: latex/figures/bar-chart-synthetic-qas.tex
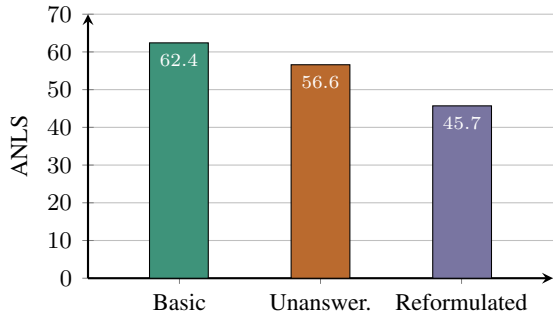
\begin{figure}[t]
\begin{tikzpicture}
\begin{axis}[
    ybar,
    axis lines=left,
    every axis x line/.append style={-Latex},
    every axis y line/.append style={-Latex},
    axis line style={thick},
    width=\columnwidth,
    height=0.66\columnwidth,
    xmin=0.5, xmax=3.5,
    ymin=0, ymax=70,
    ytick={0,10,20,30,40,50,60,70},
    ymajorgrids,
    ylabel={ANLS},
    bar width=22pt,
    xtick={1.09,2,2.91},
    xticklabels={Basic, Unanswer., Reformulated},
    tick label style={font=\small},
    label style={font=\small},
    nodes near coords,
    nodes near coords align={center},
    every node near coord/.append style={
        anchor=north,        
        font=\scriptsize,
        text=white,
        inner sep=1pt,
        yshift=-3pt
    },
    /pgf/number format/.cd, fixed, precision=1
]

\definecolor{BarTeal}{RGB}{27,158,119}
\definecolor{BarRust}{RGB}{217,95,2}
\definecolor{BarIndigo}{RGB}{117,112,179}

\addplot+[fill=BarTeal!65!gray,   draw=black] coordinates {(1.5, 62.4)};
\addplot+[fill=BarRust!65!gray,   draw=black] coordinates {(2,   56.6)};
\addplot+[fill=BarIndigo!65!gray, draw=black] coordinates {(2.5, 45.7)};

\end{axis}
\end{tikzpicture}
\caption{Extraction performance breakdown on synthetic QAs: basic vs. fully \& partially unanswerable vs. reformulated (averaged across VLMs).}
\label{fig:synthetic-breakdown}
\end{figure}

%% file: latex/sections/location_experiment.tex
\subsection{Answer Localization Results}

While ANLS measures textual extraction quality, we also assess grounding \cite{nourbakhsh-etal-2025, t-y-s-s-etal-2025-cocolex} via localization metrics: bounding-box intersection over union (IoU), normalized box proximity, and page accuracy (see Section~\ref{sec:schema_mapping}). Table~\ref{table:singlecol_model_metrics} reveals both open and closed-source models struggle with localization (maximum IoU of 14.4\%) and page identification (best accuracy of 84.3\%). Moreover, we observe that correct text predictions do not reliably entail correct spatial grounding, indicating a calibration gap between extraction and evidence localization. This difficulty may reflect the continuous, high-variance nature of bounding-box prediction and sensitivity to layout scale and spatial relations, which remain challenging for current VLMs. These limitations warrant caution in industry applications where precise localization of evidence is required, even when textual extractions are accurate.



%% file: latex/tables/localization.tex
\begin{table}[t]
\centering
\scriptsize
\setlength\tabcolsep{3pt}
\resizebox{1.0\columnwidth}{!}{
\begin{tabular}{l c c c}
    \toprule
    \textbf{Model} & \textbf{Page acc.} & \textbf{Bbox IoU} & \textbf{Bbox prox.} \\
    \midrule
    \texttt{Qwen2.5-VL-3B}      & 14.8 & 1.6 & 20.2 \\
    \texttt{Gemma-3-4B}         & 15.7 & 0.9 & 16.0 \\
    \midrule
    \texttt{Qwen2.5-VL-7B}      & 24.9 & 3.1 & 27.1 \\
    \texttt{Gemma-3-12B}        & 41.9 & 4.1 & 37.3 \\
    \texttt{Pixtral-12B}        & 39.9 & 2.5 & 38.4 \\
    \texttt{Kimi-VL-A3B-16B}    & 22.7 & 1.2 & 2.4 \\
    \midrule
    \texttt{Qwen2.5-VL-32B}     & 57.2 & \textbf{11.2} & 51.7 \\
    \texttt{Gemma-3-27B}        & 48.7 & 4.3 & 43.7 \\
    \texttt{Mistral-Small-3.2-24B} & \textbf{60.7} & 5.4 & \underline{53.2} \\
    \midrule
    \texttt{Qwen2.5-VL-72B-FP8} & \underline{59.6} & \underline{10.3} & \textbf{54.5} \\
    \midrule
    \midrule
    \rowcolor{gray!20}\texttt{Gemini-2.5-Flash}   & \underline{74.2} & \underline{13.4} & \underline{71.9} \\
    \rowcolor{gray!20}\texttt{Gemini-2.5-Pro}     & \textbf{84.3} & \textbf{14.4} & \textbf{74.8} \\
    \bottomrule
\end{tabular}
}
\caption{Model performance on page accuracy, bounding box IoU, and bounding box proximity.}
\label{table:singlecol_model_metrics}
\vspace{-1em}
\end{table}

%% file: latex/sections/structure_experiments.tex
\subsection{Output Structure Results}

\paragraph{Larger Models Yield More Valid Outputs} To ensure structured data quality, we check if (1)~model outputs are parsable in Python and if (2)~JSON-formatted leaves follow our custom extraction rules. Table~\ref{table:singlecol_model_structural} shows all models generate Python-executable structures, but only those with at least 12B parameters produce enough valid extraction leaves. \texttt{Kimi-VL-A3B-16B} is an exception, with only 5.1\% valid extraction leaves.

\paragraph{Closed Models Recall More Entities}
Table~\ref{table:singlecol_model_structural} also presents schema mapping results (see Section~\ref{sec:schema_mapping}) for each model's output. Closed VLMs extract more requested entities---\texttt{Gemini-2.5-Pro} and \texttt{Gemini-2.5-Flash} achieving recall scores of 88.1\% and 87.3\%. For open-source models, recall seems to improve with scale, suggesting that further increasing open VLM size may help cover more expected extractions and narrow the gap.


\paragraph{Structural Prediction Quality}
Tree edit similarity measures how closely the structure of a predicted output resembles that of the ground truth JSON answer. Table~\ref{table:singlecol_model_structural} shows both closed VLMs and certain large open models achieve high alignment with the ground truth. While a closed model achieves the highest scores, several open models also perform strongly, with \texttt{Gemma-3-27B} and \texttt{Qwen2.5-VL-72B} among the top performers.

%% file: latex/tables/structure.tex
\begin{table}[t]
\centering
\scriptsize
\setlength\tabcolsep{2pt}
\resizebox{\columnwidth}{!}{
\begin{tabular}{l c c c c c c}
    \toprule
    \makecell[l]{\textbf{}\\\textbf{Model}}
    & \makecell{\textbf{Valid}\\\textbf{Python}} 
    & \makecell{\textbf{Valid}\\\textbf{Ext.}} 
    & \makecell{\textbf{Map.}\\\textbf{R}} 
    & \makecell{\textbf{Map.}\\\textbf{P}} 
    & \makecell{\textbf{Map.}\\\textbf{F1}} 
    & \makecell{\textbf{Tree}\\\textbf{Sim.}} \\
    \midrule
    \texttt{Qwen2.5-VL-3B}      & 93.4 & 37.2 & 65.1 & 68.5 & 62.0 & 57.6 \\
    \texttt{Gemma-3-4B}         & 96.3 & 65.9 & 59.6 & 62.4 & 57.0 & 56.3 \\
    \midrule
    \texttt{Qwen2.5-VL-7B}      & 98.3 & 53.5 & 72.3 & 77.1 & 71.3 & 69.4 \\
    \texttt{Gemma-3-12B}        & \textbf{100.0} & 88.2 & 72.6 & 70.9 & 67.5 & 72.0 \\
    \texttt{Pixtral-12B}        & \underline{99.3} & 94.4 & 69.9 & 73.8 & 68.2 & 67.0 \\
    \texttt{Kimi-VL-A3B-16B}    & 99.0 & 5.1 & 68.3 & 68.9 & 63.4 & 57.0 \\
    \midrule
    \texttt{Qwen2.5-VL-32B}     & 98.0 & \underline{95.1} & 73.7 & 75.2 & 71.9 & 67.0 \\
    \texttt{Gemma-3-27B}        & 98.7 & 90.6 & 76.8 & 72.8 & 71.3 & \textbf{77.1} \\
    \texttt{Mistral-Small-3.2-24B} & \underline{99.3} & \textbf{99.4} & \underline{79.7} & \underline{77.3} & \underline{75.0} & 70.0 \\
    \midrule
    \texttt{Qwen2.5-VL-72B-FP8} & 99.0 & 88.8 & \textbf{80.5} & \textbf{83.8} & \textbf{79.5} & \underline{72.3} \\
    \midrule
    \midrule
    \rowcolor{gray!20}\texttt{Gemini-2.5-Flash}   & \underline{98.0} & \underline{98.6} & \underline{87.3} & \underline{81.5} & \underline{81.5} & \underline{72.6} \\
    \rowcolor{gray!20}\texttt{Gemini-2.5-Pro}     & \textbf{100.0} & \textbf{99.8} & \textbf{88.1} & \textbf{82.8} & \textbf{83.1} & \textbf{78.3} \\
    \bottomrule
\end{tabular}
}
\caption{Model perf. on format and extraction leaf validity, mapping recall/precision/F1, and tree edit similarity.}
\label{table:singlecol_model_structural}
\vspace{-1em}
\end{table}

%% file: latex/sections/answer_context_experiment.tex
\subsection{Answer Context Breakdown}

Table~\ref{table:answer-context-breakdown} in Appendix~\ref{app:answer-context-breakdown} shows that models consistently struggle in chart and free-text contexts, with 11 out of 12 models recording their lowest scores in one of these settings. Lower performance on charts was expected: charts are less structured and include visual attributes (e.g., colors, shapes) that can act as distractors, and chart parsing and understanding remain challenging for current VLMs \cite{wang2024charxiv}. As for free text, this might be due to certain queries targeting values embedded within dense pages containing substantial non-answer content and plausible near-matches, increasing the risk of false positives and hindering extraction.

%% file: latex/sections/conclusion.tex
\section{Conclusion}

We introduce \et, a benchmark dataset designed to evaluate structured IE from visually rich documents using diverse, challenging queries---including closed, schema-based, and on-demand formats---across varied document types and contexts. By expecting JSON-formatted outputs and including multi-entity, cross-page, and unanswerable queries, \et addresses key limitations of existing benchmarks and better reflects real-world extraction needs. Our experiments show that closed-source VLMs currently lead in recall and robustness, while larger models generally produce more valid outputs. However, all models face difficulties with intricate queries and answer localization, especially in less structured contexts like charts and free text. These findings highlight persistent gaps in current model capabilities. We hope this work spurs further research into more adaptable, robust, and semantically grounded extraction models for real-world applications.

%% file: latex/sections/limitations.tex
\section*{Limitations}

While \et provides a rigorous testbed for structured IE, certain limitations remain. First, our use of ANLS for text evaluation is suboptimal for certain data types, such as numbers and dates, where character-based similarity may not accurately reflect extraction quality \cite{nourbakhsh-etal-2025}; future work should explore alternative metrics tailored to these cases (26\% of all values annotated). Second, \et is currently English-only and does not assess IE performance on documents in other languages. Finally, the schema mapping stage essential to our metric computation relies on a text-only LLM to align ground truth and predicted extraction leaves. Although effective, this approach is slower than programmatic solutions and may not yield perfect mappings. Addressing these limitations in future work will further strengthen this benchmark and its utility for the broader document intelligence research.

\section*{Disclaimer}
This paper was prepared for informational purposes by the Artificial Intelligence Research group of JPMorgan Chase \& Co and its affiliates (``JP Morgan''), and is not a product of the Research Department of JP Morgan. JP Morgan makes no representation and warranty whatsoever and disclaims all liability, for the completeness, accuracy or reliability of the information contained herein. This document is not intended as investment research or investment advice, or a recommendation, offer or solicitation for the purchase or sale of any security, financial instrument, financial product or service, or to be used in any way for evaluating the merits of participating in any transaction, and shall not constitute a solicitation under any jurisdiction or to any person, if such solicitation under such jurisdiction or to such person would be unlawful.

%% file: latex/sections/appendix.tex
\clearpage

\pagebreak
\appendix

\onecolumn
\section{Data Generation Instructions}
\label{app:data-gen-instructions}
\lstset{
    breaklines=true,
    basicstyle=\fontsize{6.78}{6.75}\selectfont\ttfamily,
}
\begin{lstlisting}
# GUIDELINES
Below are guidelines to follow in order to generate query-answer pairs based on the images of an input document:
## Query:
- The "query" field of your output elements is a string.
- Queries can pertain to one of three types of extractive queries (for now, fully open information extraction queries where entire documents are requested to be parsed in a structured format are out of scope):
(a) **Closed IE with plain text:** the query directly mentions in a sentence/in the form of plain text the names of the entities to extract (ex: "what is the ID of the signers?" => [{"signer ID": "123456"}]). **These entity names mentioned in the query should not always perfectly overlap the vocabulary/language used in the document, to create some difficulty in the questions and force some semantic understanding of the entity names**! (Ex: if the document contains "study name", it might be better to ask "Give me the topic of the report" instead of something very straightforward and template-based such as "What is the name of the study?").
(b) **Closed IE with schema:** the query directly mentions in the form of a schema the names of the entities to extract (ex: "{"signer ID": ""}" => {"signer ID": "123456"}). **These entity names mentioned in the query should not always perfectly overlap the vocabulary/language used in the document, to create some difficulty in the questions and force some semantic understanding of the entity names**! (Ex: if the document contains "study name", it might be better to request "{"report_topic": ""}" instead of something very straightforward like "{"study name": ""}").
(c) **On-demand IE:** the query does not directly mention the name of all the entities to extract but instead mentions a *parent entity name* to guide the extraction (ex: "parse all the dependents' information" => [{"name": "Leo", "age": "12"}]). In this case, the answer will have to propose a schema somewhat inspired by the layout/structure of the document and populate it with values of children entities related to the parent entity requested in the query. Note that on-demand queries should remain relatively **vague** about the exact children entities that should appear in the answer (ex: "Extract all available information regarding ..."; **you should not** further specify this query by adding "..., including childrenAttribute1, childrenAttribute2, and childrenAttribute3", as these are not supposed to be known in advance). It should be up to the model tasked with answering the on-demand queries (not you) to uncover what the query exactly wants based on the parent-children implicit relationships between the entities presented in the document (ex: a person --parent entity-- can have a name, a phone number, an address, a role, etc. --children attributes--).
- Note: you can also create queries about information not present in the document, in which case the answer should contain a corresponding None value for the "text" field (see below) or should just be an empty list.
- Text queries can also be simple declarative sentences (ex: "Extract all information about ..."), besides questions.
- Try to create queries about many relevant/connected entities at the same time, and avoid lexical overlap with the document, as requested above.
## Answer:
- The "answer" field of your output elements should be a list of nested dictionaries that contain all the user-requested extractions expressed in the query, whenever the query is not a schema. If the query is an empty schema, then the answer object should just adhere to this schema and populate it with extraction dictionaries (see below).
- Extractions that compose the answer should be returned in the order of (1) appearance of the entities mentioned in the query, and (2) for multiple extractions of the same underlying type of entity, in order of appearance in the document when considering it in reading order (left to right, top to bottom).
- Dictionaries within the "answer" field can either be sub-dictionaries or sub-lists for hierarchical entities as well as for line items/templated items (ex: [{"name": "Leo", hobbies: [{"name": "guitar", "duration": "2h"}]}]), or they can be **extraction dictionaries** containing verbatim text that directly comes from the document tokens. The schema of the extraction dictionaries is: {"text": str, "page": int, "bbox": [int], "answer_context": Enum({"free_text", "templated_items", "individual_key_value", "chart", "image", "checkbox"})}. The extraction dictionaries are the final leaf nodes of the answer structure; they contain the actual text values extracted from the document. For the sake of clarity, we will replace these extraction dictionaries with their extraction text values in our examples of the "# GUIDELINES" section.
- A line item or templated item can be defined as a set of fields all referring to the same underlying parent entity (ex: if someone named X with role Y signed the document at date Z, then the corresponding line item/templated item is composed of the following child entities: {"name": X, "role": Y, "date": Z}). For example, if you generate a closed IE query that asks for the age and name of all the signers of a form, the answer could have the following structure: [{"age": W, "name": X}, {"age": Y, "name": Z}, ...]. In contrast, if the closed IE query had been "extract the document ID as well as the amount requested in the form", it does not really make sense to present the answer as [{"document ID": X, "amount requested": Y}] since "document ID" and "amount requested" are not directly related. Thus, the semantic relatedness of entities should influence the output structure (e.g., whether they are grouped in the same parent dictionary or not).
- The keys of the dictionaries that compose the answer (but are not "extraction dictionaries") should be, in order of preference (as long as a meaningful key name is proposed):
(a) the exact schema key names for closed IE user queries that directly mention a schema to populate
(b) strings that minimally overlap the user-mentioned entity names (for plain text closed queries; ex: "what is the ID of the signers?", "[{"signer ID": "123456"}]")
(c) strings coming from the document tokens (for on-demand/open queries; ex: "parse all the dependents information" => [{"name": "Leo", "age": "12"}]; also for certain plain text closed queries if it is hard to overlap an appropriate key name with the user query)
(d) strings you suggest based on your understanding of the values being extracted and your internal knowledge (for on-demand/open queries; e.g., when the string value to extract is floating text and not an actual key-value field from the doc, where the key name could be repeated verbatim; also relevant for certain closed queries if it is hard to overlap an appropriate key name with the user query).
- When you identify that the query requests line items/templated items in the answer, there is no need to name the keys of each item of the answer using XXX_1, XXX_2, XXX_3, etc; the **numeral ordering is useless**. For instance, [{"edit_1": "add"}, {"edit_2": "remove"}, {"edit_3": "modify"}] should instead be formatted as [{"edit": "add"}, {"edit": "remove"}, {"edit": "modify"}]; the numbers appended to key names are useless if the keys are returned in reading order of the doc.
- Null values VS. non-found values in the document: for closed queries (both plain text and schema-based), if one of the requested keys cannot be found, you should yield this extraction dictionary for its corresponding leaf node in the output structure: {"text": None, "page": None, "bbox": None}. If all requested keys cannot be found in the document, then you should return an empty answer list. In contrast, if one of the child entities of the parent entity requested comes from a key-value field with an empty/non-filled value in the doc (e.g., "N/A"), you should yield an extraction dictionary like the following for its corresponding leaf node: {"text": "", "page": int, "bbox": [int], "answer_context": Enum({"free_text", "templated_items", "individual_key_value", "chart", "image", "checkbox"})}}. In short, **N/As (empty/non-filled values) should be extracted as "" text values**. However, not finding requested info in a doc should yield {"text": None, "page": None, "bbox": None} extraction dictionaries or an empty list.
- Checkboxes: checkboxes can be seen as key-values too; user-requested checkboxes that are unchecked should have "No" as a value, and checked ones should have "Yes".
- For on-demand/open queries, two extraction dictionaries about the same underlying entity should not be merged together if they don't appear as one single line item/templated item in the document (ex: [{"name": "Leo S", "age": "14"}, {"name": "Leo S", "grade": "5"}] stays like that if they are present as two different line items/templated items of the same document).
## query_type: 
- string field (enum); can only be one of "closed_with_plain_text", "closed_with_schema", "on_demand".
\end{lstlisting}
\twocolumn

\section{Dataset Comparison Metrics}\label{appendix:dataset-comparison}

We compare \et with popular KEE and VQA datasets using a common set of QA-level statistics. Because KEE and VQA use different annotation formats, we harmonize them into the same view before computing metrics in Table~\ref{tab:dataset-comparison}.

\paragraph{KEE recasting}
For a KEE test set of size $n$ having an entity ontology of size $k$, we report statistics for two different recasting scenarios: 
\vspace{-0.5em}
\begin{itemize}[leftmargin=1em]
    \setlength\itemsep{-0.2em}
    \item \textbf{Document-level queries:} one query per document which always requests the same $k$ entities at once; total \#QAs $=n$. Example: \textit{``Extract the post town, postcode, and street line.''}
    \item \textbf{Entity-level queries:} $k$ queries per document, i.e. one per entity of the ontology (as in \citet{hu-2024-mplug-docowl, wang-etal-2024-docllm}); total \#QAs $=nk$. Example: \textit{``Extract the post town.''} + \textit{``Extract the postcode.''} + \textit{``Extract the street line.''}. 
\end{itemize}
\vspace{-0.5em}
Results for KEE datasets in Table~\ref{tab:dataset-comparison} show non-parenthesized values for the document-level recasting scenario and parenthesized values for the entity-level recasting scenario.

\paragraph{VQA adaptation}
We retain only \emph{extractive} QAs for the analysis of VQA test sets, i.e., QAs with all annotated answers appearing in the document text tokens. To estimate the number of entities referenced per question, a text-only model (\texttt{gpt-oss-20b}) parses entity mentions from the query. 
Example: \textit{``What is the invoice number?''} $\rightarrow$ \{\textit{``invoice number''}\};\\
\textit{``What are the subtotal and tax amounts?''} $\rightarrow$ \{\textit{``subtotal''}, \textit{``tax''}\}.

\paragraph{Statistics} We define below the metrics computed:
\vspace{-1.5em}
\begin{itemize}[leftmargin=1em]
    \setlength\itemsep{-0.2em}
    \item {\bf Number of documents (\textit{\#Docs}):} Count of documents in the test split.    
    \item {\bf Number of extractive QA pairs (\textit{\#QAs}):} Count of question-answer pairs for which the answer values can be found verbatim in the document.
    \item {\bf Average document length (\textit{Len(Doc)}):} Average word count per document.
    \item {\bf Average query length (\textit{Len(Q)}):} Average word count per query.    
    \item {\bf Average answer length (\textit{Len(A)}):} Average word count per answer (when aggregating all extracted values annotated for it).
    \item {\bf Average count of entities requested per query (\textit{\#Entities per Q}):} Average number of entity named in each query; obtained by parsing the query with an LLM for VQA datasets.
    \item {\bf Average number of extracted values in the answer (\textit{\#Values per A}):} Average count of answer values annotated for each QA.

    \item {\bf Percentage of queries requesting some missing entities (\textit{\%Q Unansw.}):} Measures how often documents lack values for at least one of the requested entities from the query.
    \item {\bf Lexical overlap (\textit{Q-Doc Overlap}):} Proportion of overlapping tokens between the query and the document text.
    \item {\bf Normalized ontology size (\textit{$\overline{|\text{Ontology}|}$}):} Ratio of the number of unique entities in the ontology to the total number of QA pairs in the test split.
\end{itemize}

\paragraph{Interpretation}
KEE datasets, when transformed with document-level queries, yield few QAs, small ontologies, and near-ubiquitous partially unanswerables (almost no document contains values for every single entity of the ontology)---all indicating limited query diversity and unrealistic coverage. When transformed with entity-level queries, KEE datasets have very short queries, single-entity requests, few answers, high query-document overlap, and an extremely low normalized ontology size due to the large number of QAs. VQA datasets instead show normalized ontology sizes near 1 (each question targets a distinct entity) but at the cost of high lexical overlap and predominantly single-entity, single-value queries. Only DUDE contains a notable fraction of fully unanswerable QAs. In contrast, \et offers a middle ground: longer multi-entity queries, many answer values expected per query, high entity diversity, a realistic proportion of queries with missing entities, and comparatively lower lexical overlap.

\section{Dataset Annotation Details}

\input{latex/figures/importance-sampling-algo}
\label{apx:dataset-annotation-details}
\paragraph{Small-Scale Manual Data Annotation}
For the manual split of \et, the validation process involved an initial review after each human annotator (four in total) had completed approximately half of their QAs---review during which a validator provided feedback and suggested adjustments based on the annotation instructions. Annotators then revised their work accordingly and completed the remaining QAs. Finally, the validator conducted a comprehensive review of all manually-annotated QAs to address any remaining issues. The validator was one of the main authors of the annotation instructions and reviewed all 102 manually generated QAs to ensure consistency.

\paragraph{Large-Scale Synthetic Data Generation} 
We generate nearly 3,000 synthetic QAs across 350 test documents from the four source datasets. We prompt \texttt{Gemini-2.5-Flash Thinking} to propose nine QAs per document (three per QA type, matching the human annotation protocol) with the instructions \ref{app:data-gen-instructions}. Then, we implement programmatic fixes to clean the QAs generated, such as JSON format repairs and bounding box reconciliation based on the OCR-processed source documents. Due to resource constraints, we could manually validate only 202 out of the nearly 3,000 synthetic QAs. We prioritize QAs for manual validation using the target-guided importance sampling algorithm detailed in Algorithm~\ref{alg:target_guided_sampling_simple}. For this, we first pre-define metadata-level desiderata (e.g., maximizing the number of requested entities or pages in the answer) for the QAs. We then sample prototypical vectors from distributions biased toward these desired characteristics and select synthetic QAs whose metadata are closest to these prototypes. This approach prioritizes QAs with target properties while avoiding overrepresentation of outliers.

After the manual validation exercise, we estimate the inter-validator agreement rate: a meta-validator reviewed a sample of 60 synthetic QAs (stratified by source dataset) out of the <202 previously validated by the other validators. For each extraction leaf, the meta-validator noted any disagreements with prior validation edits (by comparing pre- and post-validation) or identified missed changes. Agreement was calculated as 1 minus the ratio of total disagreements to total possible relevant changes across QAs. Due to resource constraints, we did not re-validate all synthetic QAs from scratch; however, the high inter-validator agreement of 96.2\% indicated no pressing need for full revalidation.

\section{Evaluation Metric for Structured IE}
\label{app:eval-metric}

\paragraph{Synthetic Dataset for Mapping Evaluation} To assess the feasibility of using an LLM ``judge'' for schema mapping in the metric computation process, we construct a controlled evaluation dataset for this intermediate task. We first curate few-shot exemplars that cover common schema variations between ground-truth and predicted structured answers (differences in hierarchy, key names, textual values, page indices, and bounding-box coordinates). Using these exemplars, we prepare two prompt templates for \texttt{Gemini-2.5-Flash}: (i)~a~transformation prompt that modifies keys, values, and permutes array orders; and (ii)~a perturbation prompt that inserts or deletes key-value pairs with probability 0.5. The resulting synthetic dataset provides systematic deviations to stress-test the mapping procedure under realistic differences from ground truth. It comprises 37 examples, all of which were manually reviewed.

\paragraph{Schema Mapping Instructions} We share the prompt fed to the text-only LLM mapper below:
\onecolumn
\lstset{
    breaklines=true,
    basicstyle=\fontsize{7.2}{7.2}\selectfont\ttfamily,
}
\begin{lstlisting}
# TASK OVERVIEW
Your are a helpful assistant that needs to match the keys of a ground truth dictionary with the keys of a prediction dictionary.
After describing the general guidelines for this task, I will show you a few examples of the matching operation your are supposed to perform; then, I will provide you with a new pair of ground truth and prediction dictionaries whose keys you will need to match.
Your output, which needs to be wrapped between "```python" and "```", will be used to evaluate the predictions of a model trained to extract structured information from documents; you will thus need to be precise in your mapping.

# GENERAL GUIDELINES
## Input characteristics:
You will receive **one ground truth dictionary** and **one prediction dictionary**:
    - **They contain values extracted from a document, tagged with semantic key names**.
    - All their keys and values are strings.
    - Both do NOT contain nested objects.
    - Each of these dictionaries was obtained by flattening an initial dictionary that contained nested objects; the key names that you will see in each of the two input dictionaries indicate the prior structure of the initial dictionaries.
        Example: the key name "0.person.age" indicates that the associated value corresponds to the age of person nb 0. If we also see the key name "0.person.address" in the same input dictionary, then it indicates that the value associated to this other key name corresponds to the address of that same person nb 0.
In addition to these two dictionaries, you will also receive the extractive query that these two dictionaries try to answer. The query is just a string.
## Matching you have to perform:
For each key-value pair of the ground truth input dictionary, you need to match the considered ground truth key-value pair to a key-value pair from the prediction dictionary. Some more instructions:
    - To match key-value pairs between the ground truth dictionary and the prediction dictionary, you should **mainly rely on the VALUE of the key-value pairs**.
    - You can rely a bit on the underlying meaning of the key strings.
    - Do not forget to pay attention to the numbers delimited by "." in the key names: they should help you identify to which parent object each children attribute belongs to.
    - **In most cases**, you should try to match one key-value pair from the ground truth dictionary to one key-value pair from the prediction dictionary. A key-value pair from the ground truth dictionary may be matched to multiple key-value pairs from the prediction dictionary iff the string value of that ground truth key can be reconstructed by concatenating/combining the string values of these multiple prediction keys.
        Example: the key "0.full name" from the ground truth dictionary {"0.full name": "John Doe", ...} should be matched to the two keys ("persons.0.First_Name", "persons.0.Last_Name") from the prediction dictionary {"persons.0.First_Name": "John", "persons.0.Last_Name": "Doe", ...}.
    - Conversely, a prediction key-value pair should **normally** be used to match only one ground truth key-value pair in your output. The only case where a prediction key-value pair may appear multiple times in your output mapping is when its string value can be reconstructed by concatenating/combining the string values of multiple ground truth key-value pairs.
        Example: the key "0.first name" from the ground truth dictionary {"0.first name": "John", "0.last name": "Doe", ...} should be matched to the key "persons.0.Full Name" from the prediction dictionary {"persons.0.Full Name": "John Doe", ...}; the key "0.last_name" from the same ground truth dictionary should also be matched to the key "persons.0.Full Name" from that prediction dictionary.
    - As most matches between the ground truth and the prediction dictionary will be one-to-one, you should consider the overall effect of each individual match on the rest of the mapping. Do not focus only on greedily making the "best" local matches for a few ground truth keys at the expense of others (indeed, most prediction keys will not be eligible for reuse). Instead, proceed holistically to return a mapping where all ground truth keys are matched as well as possible.
    - Try to match as many key-value pairs from the ground truth dictionary to key-value pairs from the prediction dictionary. However, **if a key-value pair from the ground truth dictionary has truly no equivalent(s) in the prediction dictionary, please LEAVE IT UNMATCHED** and put None in the corresponding map (e.g., if "person.age" from the ground truth dictionary cannot be matched to a key-value pair from the prediction dictionary, then your output dictionary should contain {..., "person.age": None}).
## Output format expected
Your final output should be a python dictionary wrapped between "```python" and "```" that maps all the string key names of the ground truth dictionary to string key names from the prediction dictionary. Since it can happen that a ground truth key is matched to multiple prediction keys, all the values of your output python dictionary should be **tuples of key names from the ground truth dictionary** (most will be singletons). If no match with a prediction key-value pair is found, the value should be None.

# FEW SHOT EXAMPLES
Here are a few examples of what you are supposed to do given a ground truth dictionary, a prediction dictionary, and a query --along with explanations of why the proposed fuzzy mapping is the way it is.
## Example nb 1:
### Ground truth dictionary:
```python
{'0.start_date': '01/01/94', '0.end_date': '12/31/94', '0.TOTAL COST': '$210,910', '1.start_date': '01/01/95', '1.end_date': '12/31/95', '1.TOTAL COST': '$212,481', '2.start_date': '01/01/96', '2.end_date': '12/31/96', '2.TOTAL COST': '$220,416'}
```
### Prediction dictionary:
```python
{'slot1.expenses': '212481', 'slot1.period start': '01-01-95', 'slot1.period end': '12-31-95', 'slot2.dt_begin': '01-01-94', 'slot2.dt_final': '12-31-94', 'slot3.price': '220416' 'slot3.start': '01 01 96', 'slot3.end': '12 31 96'}
```
### Corresponding extractive query:
"What was the cost over time of the project?"
### Proposed output fuzzy mapping:
```python
{'0.start_date': ('slot2.dt_begin',), '0.end_date': ('slot2.dt_final',), '0.TOTAL COST': None, '1.start_date': ('slot1.period start',), '1.end_date': ('slot1.period end',), '1.TOTAL COST': ('slot1.expenses',), '2.start_date': ('slot3.start',), '2.end_date': ('slot3.end',), '2.TOTAL COST': ('slot3.price',)
}
```
Explanation (just so that you understand some important details): here, we could NOT match '0.TOTAL COST' to any key from the prediction dictionary because, although item 0 from the ground truth does match slot2 from the prediction dictionary, no sub-key of slot2 provides a cost value. If slot2 had had a sub-key entitled "cost" or something along these lines, even with a None or "" or "N/A" value, we could have matched "0.TOTAL COST" to "slot2.cost".
## Example nb 2:
### Ground truth dictionary:
```python
{'0.percentage_of_prime_working_age_women_discovering_car_manufacturers_through_magazines': '17%'}
```
### Prediction dictionary:
```python
{'0.value': '21', '0.unit': '%'}
```
### Corresponding extractive query:
"What percentage of prime working-age women discover car manufacturers through magazines?"
### Proposed fuzzy mapping:
```python
{'0.percentage_of_prime_working_age_women_discovering_car_manufacturers_through_magazines': ('0.value', '0.unit')}
```
Explanation (just so that you understand some important details): here, even though the values are different for the single key of the ground truth and for the keys of the prediction dictionaries ("17%" VS. "21" + "%"), these values seem to refer to the same underlying entity --based on how the keys are named and on the fact that we are requesting **a single** type of percentage in the query. Indeed, here, even though the extracted values in the prediction dictionary are incorrect ("21" + "%" VS. "17%"), the prediction does seem to have attempted to extract the "percentage of prime working-age women discover car manufacturers through magazines". It is just that it called it "0.value" and "0.unit" instead of "0.percentage_of_prime_working_age_women_discovering_car_manufacturers_through_magazines". Also, as you can see, the prediction dictionary split the requested percentage value into two key-value pairs ({"0.value": "21", "0.unit": "%"}), whereas the ground truth dictionary combined both into the a single key-value pair. We thus had to match the single key-value pair from the ground truth to the two key-value pairs from the prediction dictionary. Had the situation been reversed (i.e., two more granular key-value pairs in the ground truth corresponding to one coarser key-value pair in the prediction), we would have had to match each key-value pairs from the ground truth dictionary to the same prediction key-value pair, exceptionally.
## Example nb 3:
### Ground truth dictionary:
```python
{'0.name': 'John Doe Alpha', '0.position': 'President & CEO', '0.dept': None, '0.phone': '212-555-0101', '0.email': 'alpha@doe-bureau.example.com', '1.name': 'John Doe Beta', '1.position': 'EVP', '1.dept': None, '1.phone': '212-555-0102', '1.email': 'beta@doe-bureau.example.com', '2.name': 'John Doe Gamma', '2.position': 'SVP', '2.dept': 'Strategic Sales Insights', '2.phone': '212-555-0103', '2.email': 'gamma@doe-bureau.example.com', '3.name': 'John Doe Delta', '3.position': 'VP', '3.dept': 'Strategic Insights', '3.phone': '212-555-0104', '3.email': 'delta@doe-bureau.example.com', '4.name': 'John Doe Epsilon', '4.position': 'Sr Director', '4.dept': 'Strategic Research Insights', '4.phone': '212-555-0105', '4.email': 'epsilon@doe-bureau.example.com'}
```
### Prediction dictionary:
```python
{'authors.0.name': 'John Doe Alpha', 'authors.0.title': 'President & CEO', 'authors.0.email': 'alpha@doe-advertising.example.com', 'authors.1.name': 'John Doe Betta', 'authors.1.title': 'EVP', 'authors.1.email': 'beta@doe-advertising.example.com', 'authors.2.name': 'John Doe Gamme', 'authors.2.title': 'SVP Strategic Sales Insights', 'authors.2.email': 'gamma@doe-advertising.example.com', 'authors.3.name': 'John Doe Epsilon', 'authors.3.title': 'Sr Director Strategic Research Insights', 'authors.3.email': 'epsilon@doe-advertising.example.com', 'authors.4.name': 'John Doe Delta', 'authors.4.title': 'VP Strategic Insights', 'authors.4.email': 'delta@doe-advertising.example.com'}
```
### Corresponding extractive query:
"Who are the authors of the slides? Give all information about them."
### Proposed fuzzy mapping:
```python
{'0.name': ('authors.0.name',), '0.position': ('authors.0.title',), '0.dept': None, '0.phone': None, '0.email': ('authors.0.email',), '1.name': ('authors.1.name',), '1.position': ('authors.1.title',), '1.dept': None, '1.phone': None, '1.email': ('authors.1.email',), '2.name': ('authors.2.name',), '2.position': ('authors.2.title',), '2.dept': ('authors.2.title',), '2.phone': None, '2.email': ('authors.2.email',), '3.name': ('authors.4.name',), '3.position': ('authors.4.title',), '3.dept': ('authors.4.title',), '3.phone': None, '3.email': ('authors.4.email',), '4.name': ('authors.3.name',), '4.position': ('authors.3.title',), '4.dept': ('authors.3.title',), '4.phone': None, '4.email': ('authors.3.email',)}
```
Explanation (just so that you understand some important details): here, even though the values extracted in the predictions are not all perfectly accurate compared to the ground truth overall, it is still easy to map ground truth keys to prediction ones based on their names/structure and how similar their values are. It is also clear that the prediction dictionary forgot all the phone number extractions, and that the prediction key-value pairs corresponding to "John Doe Delta" have index "3" VS. index "4" in the ground truth dictionary (this really underlines that, even if the key names are very similar/share the same index between gt and pred, they are not necessarily the best match: their corresponding gt and pred values should help you determine which gt and pred keys go together). Finally, we note that the ground truth dictionary was more granular regarding "position" and "dept" information (separate key-value pairs for each, per author), whereas the prediction dictionary grouped both of these pieces of information into the same key-value pair "title" when department information is available, for each author. Thus, the "authors.2.title", "authors.3.title", and "authors.4.title" prediction keys can be used twice in the output dictionary, since their values contain both "position" and "dept" information. As for "authors.0.title" and "authors.1.title", since their values do not contain any department information, the ground truth keys "0.dept" and "1.dept" should not be matched to them (as opposed to "0.position" and "1.position"), hence their None values in the output dictionary.

# YOUR TURN
Now, you will be provided with a new ground truth dictionary, a new prediction dictionary, and a new extractive query. After some reasoning, you will need to propose a fuzzy mapping between the key names of the ground truth dictionary and of the prediction dictionary. Remember: your final fuzzy mapping should be wrapped between "```python" and "```".
## Ground truth dictionary:
```python
\end{lstlisting}
\twocolumn
If the LLM mapper does not include all ground truth keys in its output mapping, we re-prompt it in the conversation to map the missing ground truth key-value pairs to predicted ones. The same correction process is applied if the LLM mapper matches a ground truth key to a key that cannot be found among the predicted key-value pairs. Finally, we club ground truth keys in the output mapping in order to align the two schemas to the same level of granularity.

\section{Text Extraction Results: Breakdown by Answer Context}
\label{app:answer-context-breakdown}
\input{latex/tables/anls_answer_context_breakdown}

\section{VLM Benchmarking Prompt}
The prompt used to test VLMs on \et can be found in the next page. The three shots (one per QA type, based on FUNSD source documents for brevity) from the manually-annotated subset of \et are removed for space considerations.
\newpage
\onecolumn
\lstset{
    breaklines=true,
    basicstyle=\fontsize{6.6}{6.5}\selectfont\ttfamily,
}
\begin{lstlisting}
# OVERVIEW
You are a helpful assistant required to answer **extractive** queries based on document images that the user will provide. 
You should always return your final answer in a structured object format presented as a list of (potentially-nested) dictionaries or a populated schema adhering to that of the input query.
NOTE: extractive means that the answer must be contained in the document; in the rest of the instructions, "IE" stands for "Information Extraction".
# GUIDELINES
Below are guidelines to follow in order to correctly answer queries based on the images of an input document (i.e., extract the right information, using the right output structure):
## Input query:
- The input "query" you will receive is a string.
- Queries can be formulated either in plain text or as "empty" structured object schemas that you need to populate with extracted values.
- Queries can pertain to one of three types of extractive queries:
    (a) **Closed IE with plain text:** the query directly mentions in a sentence/in the form of plain text the names of the entities to extract (ex: "what is the ID of the signers?" => [{"signer ID": "123456"}])
    (b) **Closed IE with schema:** the query directly mentions in the form of a schema the names of the entities to extract (ex: "{"signer ID": ""}" => {"signer ID": "123456"})
    (c) **On-demand IE:** the query does not directly mention the name of all the entities to extract but instead mentions a *parent entity name* to guide the extraction (ex: "parse all the dependents information" => [{"name": "Leo", "age": "12"}]). In this case, your answer will have to propose a schema somewhat inspired by the layout/structure of the document and populate it with extracted values of children entities related to the parent entity requested in the query.
- Queries can be simple declarative sentences (ex: "Extract all information about ..."), besides questions.
## Your output answer:
- Your answer should be a list of nested dictionaries that contain (inside extraction dictionaries) all the user-requested extractions expressed in the query, **whenever the query is not a schema**. When the query is an empty schema, then the answer object should just adhere to this schema and populate it with extraction dictionaries (see below).
- The extractions that compose your answer should be returned in order of (1) appearance of the entities mentioned in the query, and (2) for multiple extractions of the same underlying type of entity, in order of appearance in the document when considering it in reading order (left to right, top to bottom).
- Dictionaries within your structured answer can either be nested dictionaries/sub-lists when extracting hierarchical entities or line items/templated items (ex: [{"name": "Leo", hobbies: [{"name": "guitar", "duration": "2h"}]}]), OR **extraction dictionaries** containing verbatim text directly extracted from the document tokens. The schema of the extraction dictionaries that will be the leaf nodes of your output structure is: {"text": str, "page": int, "bbox": [int]}. The extraction dictionaries are the final/leaf nodes of the structured output that contain the actual text values you extract from the document. For the sake of clarity, we replace these extraction dictionaries with the text values in our examples of the # GUIDELINES section.
- Within an **Extraction dictionary** (i.e., a leaf node of your output structure), the page field should be an integer referring to the index of the page where the extracted text value came from. The text "### New doc image nb i" will provide page indices in the user prompt (0-indexed).
- Within an **Extraction dictionary**, the bounding box coordinates (see "bbox" field) should be normalized between 0 and 1000 and should correspond to the bounding box of the rectangle containing the extracted text.
- A line item or templated item can be defined as a set of fields all referring to the same underlying parent entity (ex: if someone named X with role Y signed the document at date Z, then the line item/templated item is composed of the following child entities: {"name": X, "role": Y, "date": Z}). This is particularly important for correctly structuring your answer object. For example, if you come across a closed IE query that asks for the age and name of all the signers of a form, and that all of this information is presented in a templated manner, your answer should have the following structure: [{"age": W, "name": X}, {"age": Y, "name": Z}, ...]. In contrast, if the closed IE query had been "extract the document ID as well as the amount requested in the form", it does not really make sense to present the answer as [{"document ID": X, "amount requested": Y}] since "document ID" and "amount requested" are not directly related (both semantically and from a layout perspective).
- The key names of the dictionaries that compose your answer (but that are not leaf "extraction dictionaries") should be, in order of preference (as long as a meaningful key name is proposed):
    (a) the actual schema key names for closed IE user queries that directly mention a schema to populate (you just need to adhere to the input schema and fill it with extraction dictionaries)
    (b) strings that minimally overlap the user-mentioned entity names (for plain text closed queries; ex: "what is the ID of the signers?", "[{"signer ID": "123456"}]")
    (c) strings coming from the document tokens (for on-demand queries; ex: "parse all the dependents information" => [{"name": "Leo", "age": "12"}]; also for certain closed queries if it is hard to overlap an appropriate key name with the user query)
    (d) strings you suggest based on your understanding of the values being extracted and your internal knowledge (for on-demand queries; ex: when the string to extract is floating text, not an actual key-value where the key name could be repeated; also for certain plain text closed queries if it is hard to overlap an appropriate key name with the user query).
- When you identify that the query requests line items/templated items in the answer, there is no need to name the keys of each item of the answer using XXX_1, XXX_2, XXX_3, etc; the **numeral ordering is useless** (ex: [{"edit_1": "add"}, {"edit_2": "remove"}, {"edit_3": "modify"}] should instead be formatted as [{"edit": "add"}, {"edit": "remove"}, {"edit": "modify"}]; the numbers appended to key names are useless if the keys are returned in reading order of the doc).
- Null values VS. non-found values in the document: for closed queries (both plain text and schema-based), if one of the requested keys cannot be found, you should yield this extraction dictionary for its corresponding leaf node in the output structure: {"text": None, "page": None, "bbox": None}. If all requested keys cannot be found in the document, then you should return an empty answer list. In contrast, if one of the child entities of the parent entity requested comes from a key-value field with an empty/non-filled value in the doc (e.g., "N/A"), you should yield an extraction dictionary like the following for its corresponding leaf node: {"text": "", "page": int, "bbox": [int]}. In short, **N/As (empty/non-filled values) should be extracted as "" text values**. However, not finding requested info in a doc should yield {"text": None, "page": None, "bbox": None} extraction dictionaries or an empty list.
- Checkboxes: checkboxes can be seen as key-values too; user-requested checkboxes that are unchecked should have "No" as text value, and checked ones should have "Yes".
- For on-demand queries, two extraction dictionaries about the same underlying entity should not be merged together if they don't appear as one single line item/templated item in the document (ex: [{"name": "Leo S", "age": "14"}, {"name": "Leo S", "grade": "5"}] stays like that if they are present as two different line items/templated items of the same document).
- For closed queries where the user inputs an empty schema that you are expected to populate, your structured answer should indeed populate the user schema but it should have **extraction dictionaries** as leaf nodes, still. In other words, do not only populate input user schemas with text values: populate them with extraction dictionaries/leaf nodes that contain "text", "page", and "bbox" key-value pairs.
- Please wrap your structured answer containing the extraction dictionaries between "```python" and "```".
# FEW SHOT EXAMPLE nb ...
Below are document images + an example query-answer pair to show you more concretely the type of answer you should be generating given an input query.
## Document images: [...]
## Input query for this document: "[...]"
## Output answer to the above query: ```python[...]```
# YOUR TURN NOW
Below are images of a new document. Please provide a structured answer to the query that will be given at the end, by taking into account both the initial guidelines and the example(s) provided above, if any. Make sure to wrap your structured answer between "```python```" and "```".
## New document images: [...]
## Input query for this new document: "[...]"
## Your output answer to the above query for the new document:
\end{lstlisting}

%% file: latex/figures/importance-sampling-algo.tex
\begin{algorithm*}[h]
\small
\caption{Target-Guided Importance Sampling in Feature Space}
\label{alg:target_guided_sampling_simple}
\SetKwInput{KwIn}{Input}
\SetKwInput{KwOut}{Output}
\KwIn{Dataset $\mathcal{D} = \{x_i\}_{i=1}^N$ with $x_i \in \mathbb{R}^k$ (metadata vector of QA $i$); desiderata $\mathcal{S} = \{(d_j, w_j)\}_{j=1}^k$ with $d_j \in \{\text{maximize}, \text{minimize}\}$; number of targets $M$; number of samples $n_{\text{pick}}$; $d(\cdot,\cdot)$ a distance on $[0,1]^k$ (e.g., Mahalanobis in percentile space with regularization).}
\KwOut{Subsampled set $\mathcal{D}_{\text{sub}} \subset \mathcal{D}$.}

\textbf{Step 1: Normalize features between 0 and 1}\;
Compute all normalized features $R \in [0,1]^{N \times k}$ for $\mathcal{D}$ (e.g., percentile-normalization to downweight outliers)\;

\textbf{Step 2: Sample target vectors}\;
\For{$m \gets 1$ \KwTo $M$}{
  \For{$j \gets 1$ \KwTo $k$}{
    \eIf{$d_j = \text{maximize}$}{
      $t_{m,j} \sim \mathrm{Beta}(\alpha_j, 1)$ where $\alpha_j$ increases with $w_j$\;
    }{
      $t_{m,j} \sim \mathrm{Beta}(1, \beta_j)$ where $\beta_j$ increases with $w_j$\;
    }
  }
}

\textbf{Step 3: Nearest-neighbor assignment (without replacement)}\;
$\mathcal{I} \gets \{1,\dots,N\}$, $\mathcal{D}_{\text{sub}} \gets \emptyset$\;
\For{$m \gets 1$ \KwTo $M$}{
  \If{$|\mathcal{D}_{\text{sub}}| = n_{\text{pick}}$}{\textbf{break}\;}
  $i^* \gets \arg\min\limits_{i \in \mathcal{I}} d\!\left(R_i,\, t_m\right)$\;
  Add $x_{i^*}$ to $\mathcal{D}_{\text{sub}}$ and remove $i^*$ from $\mathcal{I}$\;
}
\Return{$\mathcal{D}_{\text{sub}}$}\;
\end{algorithm*}

%% file: latex/tables/anls_answer_context_breakdown.tex
\begin{table}[h!]
\centering
\scriptsize
\setlength\tabcolsep{2pt}
\resizebox{\columnwidth}{!}{
\begin{tabular}{l c c c c c}
    \toprule
    \makecell[l]{\textbf{}\\\textbf{Model}}
    & \makecell{\textbf{Free}\\\textbf{Text}} 
    & \makecell{\textbf{Key-Value}\\\textbf{Field}} 
    & \makecell{\textbf{Templated}\\\textbf{Items}} 
    & \makecell{\textbf{Chart}} 
    & \makecell{\textbf{Checkbox}} \\
    \midrule
    \texttt{Qwen2.5-VL-3B}      & 39.9 & 52.6 & 43.6 & 28.3 & 38.4 \\
    \texttt{Gemma-3-4B}         & 29.2 & 51.8 & 23.5 & 16.4 & 36.0 \\
    \midrule
    \texttt{Qwen2.5-VL-7B}      & 48.2 & 59.8 & 55.3 & 38.3 & 58.3 \\
    \texttt{Gemma-3-12B}        & 37.6 & 66.8 & 51.0 & 26.7 & 62.4 \\
    \texttt{Pixtral-12B}        & 31.1 & 55.1 & 53.1 & 37.3 & 31.5 \\
    \texttt{Kimi-VL-A3B-16B}    & 37.1 & 63.8 & 47.7 & 26.5 & 35.8 \\
    \midrule
    \texttt{Qwen2.5-VL-32B}     & \underline{56.2} & \textbf{77.2} & 58.0 & \underline{40.5} & 57.6 \\
    \texttt{Gemma-3-27B}        & 38.2 & \underline{76.4} & 65.4 & 38.5 & 29.7 \\
    \texttt{Mistral-Small-3.2-24B} & 37.1 & 75.1 & \textbf{69.6} & \textbf{45.9} & \underline{72.3} \\
    \midrule
    \texttt{Qwen2.5-VL-32B-FP8} & \textbf{58.3} & 76.2 & \underline{69.4} & 39.6 & \textbf{81.2} \\
    \midrule
    \midrule
    \rowcolor{gray!20}\texttt{Gemini-2.5-Flash}   & \underline{69.5} & \underline{91.1} & \underline{78.1} & \underline{69.0} & \underline{84.5} \\
    \rowcolor{gray!20}\texttt{Gemini-2.5-Pro}     & \textbf{77.4} & \textbf{92.2} & \textbf{85.5} & \textbf{76.9} & \textbf{86.8} \\
    \bottomrule
\end{tabular}
}
\caption{Model performance on free text, key-value field, templated items (which include table rows), chart, and checkbox formats.}
\label{table:answer-context-breakdown}
\vspace{-1em}
\end{table}